%% file: main.tex
\pdfoutput=1

\documentclass[11pt]{article}
\usepackage[dvipsnames]{xcolor}
\usepackage[]{acl}

\usepackage{times}
\usepackage{latexsym}

\usepackage[T1]{fontenc}

\usepackage[utf8]{inputenc}
\usepackage[colorinlistoftodos,prependcaption,textsize=tiny]{todonotes}

\usepackage{microtype}

\usepackage{hyperref}
\usepackage{booktabs}
\usepackage{graphicx}
\graphicspath{{./imgs/}}
\usepackage{footmisc}
\usepackage{soul}

\usepackage{microtype}

\usepackage{caption}
\usepackage{subcaption}

\usepackage{tabularx,colortbl}
\definecolor{gray}{gray}{0.85}
\usepackage[dvipsnames]{xcolor}
\usepackage{tikz}
\usepackage{collcell}

\input{math-com}
\usepackage[capitalize,noabbrev]{cleveref}

\definecolor{applegreen}{rgb}{0.55, 0.71, 0.0}
\definecolor{cadmiumred}{rgb}{0.89, 0.0, 0.13}
\definecolor{azure}{rgb}{0.0, 0.5, 1.0}
\definecolor{amber}{rgb}{1.0, 0.75, 0.0}
\definecolor{darkorchid}{rgb}{0.6, 0.2, 0.8}

 \newcommand{\change}[1]{{\leavevmode\color{black}#1}}

\title{OpenCQA: Open-ended Question Answering with Charts}

\author{
Shankar Kantharaj$^{1}$, Xuan Long Do$^{2}$, Rixie Tiffany Ko Leong$^{2}$,\\  \textbf{Jia Qing Tan}$^{2}$\textbf{,} \textbf{Enamul Hoque$^{1}$}\textbf{,} \textbf{Shafiq Joty$^{2,3}$}\\
$^1$York University, Canada, $^2$Nanyang Technological University, Singapore, \\
$^3$Salesforce AI Research \\
\texttt{\{shankark, enamulh\}@yorku.ca}\\
\texttt{\{xuanlong001@e, C190022@e, srjoty@\}.ntu.edu.sg} \\
}
\begin{document}
\maketitle

\begin{abstract}
Charts are very popular to analyze data and convey important insights. People often analyze visualizations to answer open-ended questions that require explanatory answers. Answering such questions are often difficult and time-consuming as it requires a lot of cognitive and perceptual efforts. To address this challenge, we introduce a new task called OpenCQA, where the goal is to answer an open-ended question about a chart with descriptive texts. We present the annotation process and an in-depth analysis of our dataset. We implement and evaluate a set of baselines under {three practical}  settings. \change{In the first setting, a chart and the accompanying article 
is provided as input to the model. 
The second setting provides only the relevant paragraph(s) to the chart instead of the entire article, whereas the third setting requires the model to generate an answer solely based on the chart.}
Our analysis of the results show that the top performing models generally produce fluent and coherent text while they struggle to perform complex logical and arithmetic reasoning. 

\end{abstract}

\section{Introduction}

\begin{figure}[t!]
\centering
\includegraphics[width=0.44\textwidth, trim={0cm 0cm 0cm 0cm},clip]{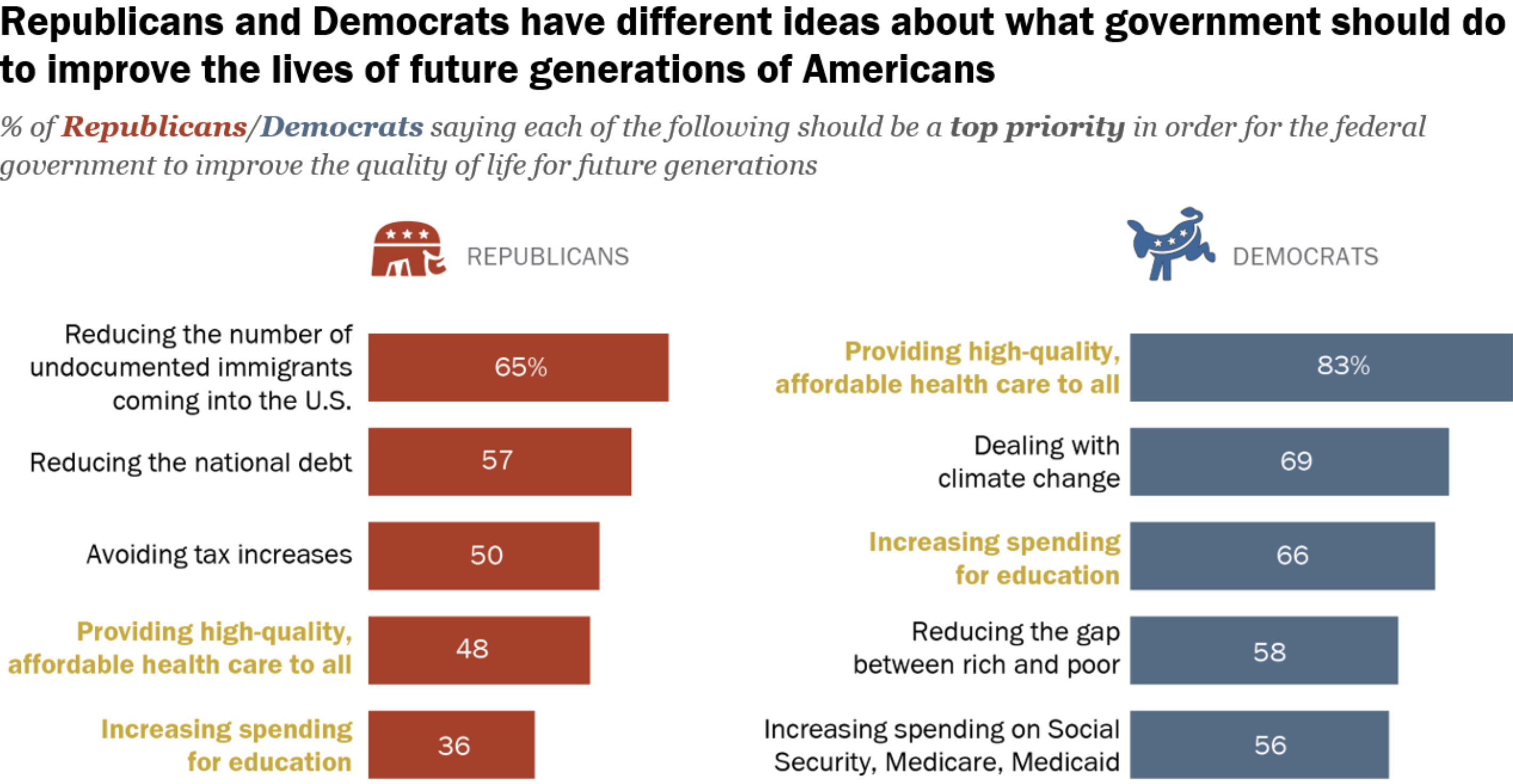}
\vspace{1mm}
{\begin{tabular}{p{7.2cm}}
\small \underline{\textbf{Question}}: Compare the Democrats and Republicans views about providing health care to the population?\\ 
\small \underline{\textbf{Answer }}: While 83\% of Democrats say providing high-quality, affordable health care for all should be a top priority, a much smaller share of Republicans (48\%) agree.
\end{tabular}}
\vspace{-3mm}
\caption{\small A question-answer pair from our dataset.}
\vspace{-5mm}
\label{fig:teaser}
\end{figure}

Using data visualizations 
such as bar charts and line charts to discover critical insights, and explain them to others is at the heart of many decision making tasks~\cite{munzner2014visualization}. Often, people explore such visualizations to answer  high-level questions that involve reasoning and explanations.
For example, \Cref{fig:teaser} shows an open-ended question which cannot be answered by a single word or phrase, rather it requires an explanatory answer. Answering such questions can be time consuming and mentally taxing as they require significant amounts of perceptual and cognitive efforts. For the particular question in \Cref{fig:teaser}, the user needs to find relevant marks (bars) in the given charts, compare their values and perform  reasoning over them to generate an explanatory answer. \change{Thus, the research question we address in this paper is: can we build systems to} automatically answer such open-ended questions about charts with descriptive texts?

Chart Question Answering (CQA) is a task where the goal is to take a chart and a natural language question as input and generate the desired answer as output~\cite{hoque-etal-survey}. While CQA has received growing attentions in the last few years, existing datasets only focus on close-ended (factoid) questions where the answer is a word or phrase~\cite{kahou2017figureqa, dvqa, leafqa, stlcqa}. These datasets typically use predefined templates to generate synthetic questions {and answers to these questions come from a closed vocabulary (\eg\ `yes', `no', `x-axis-label'). 
PlotQA  \cite{plotqa} introduces some open-vocabulary questions that require aggregation operations on the underlying chart data; the answer is still however a number or a word/phrase obtained from the chart. \citet{kim2020answering} attempt to automatically explain how the model computes the answer but only for close-ended ones. To our knowledge, there are no datasets on CQA with open-ended questions. 

In this work, we introduce a novel task named, OpenCQA in which the system takes a chart and a question as input and is expected to produce a descriptive answer as output like the one in \Cref{fig:teaser}. This task is multidisciplinary and challenging in nature as it involves natural language generation (NLG), information visualization and computer vision. It differs from the data-to-text and reading comprehension, because unlike text or tables, charts serve a different communicative goal by creating visual representation of data. Readers can quickly notice important patterns, trends, and outliers from such visual representation which cannot be easily observed from a table of raw data~\cite{munzner2014visualization}. By looking at a line chart, one can quickly discern an important trend whereas scatterplots may visually depict correlations and outliers.  Existing NLG approaches for 
tables do not consider such chart features in the generation.


We have developed a benchmark dataset {for OpenCQA} consisting of 7,724 human-written open-ended questions about a variety of real-world charts and the associated descriptive answers. \change{We formulate three practical task settings. In the first setting, a chart and the article containing the chart is provided as input, and the model generates an answer to an open-ended question. This setting poses extra challenge as articles often contain paragraphs that are irrelevant to the questions. 
To make the task more focused, the second setting provides only the relevant paragraph(s) to the chart. {Hence we can measure the models' ability to answer a question without the effect of extra noise from irrelevant text.}
The third setting is more challenging as the related text is not provided and the model needs to generate an answer solely based on the chart. This is more relevant to real world scenarios where charts are not associated with any explanatory text.} 
}

Since the proposed task is completely new, we adapt a variety of state-of-the-art models that utilize \change{multimodal}, data2text and extractive summarization methods to serve as strong baselines.
\change{
We conducted automatic and qualitative evaluations and observe that the top performing models are quite fluent and coherent in generating summaries but lack in complex logical reasoning and inference. Our codebase is publicly available at \url{https://github.com/vis-nlp/OpenCQA}.} 

\section{Related Work}
\vspace{-0.2em}

Our work is related to {three} lines of prior work. 

\vspace{-0.6em}
\paragraph{(i) Chart Summarization}
 \citet{mittal-etal-1998-describing} and \citet{Ferres-accessibility-2013} adopt a planning-based architecture and used templates to describe charts with texts. These methods can only describe how to read a chart without summarizing any insights from the chart. \citet{demir-etal-2012-summarizing} compute statistics to generate bar chart summaries and simultaneously construct sentence- and discourse-level structures.  \citet{DBLP:journals/corr/abs-1906-02850} use a ResNet \cite{he2016deep} to encode a chart  and a LSTM decoder to generate a caption. All these studies generate summaries using predefined templates, which may lack naturalness and variations in terms of grammatical structure and lexical choices. \citet{obeid} and \citet{kantharaj-etal-2022-chart} use transformer-based models  while \citet{Andrea-carenini} use an LSTM based encoder-decoder model to generate chart summaries in a data-driven fashion. But their models only focus on generating a summary to describe the chart rather than focusing on a specific relevant portion of a chart to answer a question which is the main focus of our work.

\vspace{-0.6em}
\paragraph{(ii) Visual Question Answering (VQA)} \change{VQA involves answering a question regarding an input image~\cite{antol2015vqa}.
To relate the question and the image effectively,  researchers focus on fusing textual and visual information together \citep{lu2019vilbert,talmor2021multimodalqa}. \citet{cho2021unifying} introduce VL-T5 and VL-BART 
as pretrained vision-language models which achieved competitive results on VQA tasks. 
Unlike images with real-world objects and scenes, charts encode data using marks (bars, lines) and have inherent structure which makes the chart QA task quite different from VQA \cite{masry-etal-2022-chartqa}.
}

\vspace{-0.5em}
\paragraph{(iii) Data2text Generation} Data2text models generate a descriptive summary from a data table. Previous work has focused on specific domains such as  sports~\cite{barzilay-lapata-2005-collective,  
wiseman-etal-2017-challenges}, weather-forecast~\cite{reiter2005choosing}, recipe \cite{yang2017reference} and biography \cite{lebret-etal-2016-neural}. Others~\cite{parikh2020totto, chen2020logical} have focused on  open-domain tasks. Many of these methods use an LSTM-based encoder-decoder architecture~\cite{mei2016talk, lebret-etal-2016-neural, wiseman-etal-2017-challenges}, while  \citet{gong-etal-2019-enhanced} find that transformers yield more fluent and coherent outputs. Few approaches focus on generating textual facts with logical inference rather than stating simple facts that can be easily retrieved from the data table~\cite{chen2020logical,chen-etal-2020-logic2text}. Unlike the task with data tables, our task involves understanding visual features of the charts and the natural language questions to perform reasoning in order to generate (or extract) texts as answers.

\section{Dataset Construction}

\input{Dataset}

\section{OpenCQA Models} \label{sec:model}
\input{Models}
\section{Evaluation}


\input{Evaluation}
\vspace{-2mm}
\section{Conclusion}
\vspace{-0.5em}
We propose a new benchmark, OpenCQA, for answering open-ended questions about charts using descriptive texts. We also introduce several state-of-the-art baselines and measures. Our evaluation 
suggests that current state of the art generative models can produce fluent text but still 
struggle to produce 
relevant and factually correct statements with numerical and logical reasoning. We hope that OpenCQA will serve as a useful research benchmark for model and metric development and
motivate other researchers to explore this new task.

\section*{Ethical Considerations}
During the dataset collection and annotation process, we had many ethical issues to take into consideration. To respect the intellectual property of the chart publishers, we only used publicly available charts from resources that provide publication rights of downloaded content for academic purposes. According to the terms and conditions for Pew,\footnote{\href{https://www.pewresearch.org/about/terms-and-conditions/}{https://www.pewresearch.org/about/terms-and-conditions/}} users are allowed to use the content as long as they are attributed to the Center or are not attributed to a different party. 

To fairly compensate the Mechanical Turk annotators, we compensated the annotators based on the minimum wage in the US (7.25 US\$ per hour), which is 1 US\$ for each task.  
Additionally, to protect the privacy of these annotators, all of their annotations were anonymized.  To ensure the reproducibility of our experimental results, we have provided the hyperparameter settings 
in \Cref{app:baselines}.

We foresee one possible misuse of our models that is to spread misinformation. Currently, our model outputs tend to appear fluent but contain some factual errors, as detailed in \Cref{sec:error-analysis}. Hence, if such model outputs are published without being corrected, it may mislead and misinform the general public.

\section*{Limitations}
A limitation of our dataset is that we were limited to Pew research (pewresearch.org) as a source due to the nature of our task and thus could not consider other sources. Future work could expand on our dataset when suitable sources become available. Further, we did not consider long-range sequence models such as linformer \cite{wang2020linformer} or recently proposed memorized transformer \cite{wu2022memorizing} for modeling long sequences. Lastly, our task setup is limited since we only have access to the automatically extracted OCR data which is often noisy. Future methods can focus on improving OCR extraction for this task to improve the input of the model.

\section*{Acknowledgement}
The authors would like to thank the anonymous reviewers for their helpful comments. This research was supported by the Natural Sciences \& Engineering Research Council (NSERC) of Canada.

\bibliographystyle{acl_natbib}
\bibliography{main}
\newpage
\input{Appendix}

\end{document}

%% file: math-com.tex
\usepackage{amsmath}
\usepackage{amsfonts,bm}
\usepackage{xspace}

\newcommand{\red}[1]{\textcolor{red}{#1}}

\newcommand{\ie}{{\em i.e.,}\xspace}
\newcommand{\eg}{{\em e.g.,}\xspace}
\newcommand{\wrt}{\emph{w.r.t.}\xspace}

\newcommand{\Ni}{({\em i})~}
\newcommand{\Nii}{({\em ii})~}
\newcommand{\Niii}{({\em iii})~}
\newcommand{\Niv}{({\em iv})~}

\definecolor{mypink3}{cmyk}{0, 0.7808, 0.4429, 0.1412}

\makeatletter   
\newcommand{\sveryshortarrow}[1][3pt]{\mathrel{%
    \vcenter{\hbox{\rule[-.5\fontdimen8\scriptfont3]
               {\scriptratio\dimexpr#1\relax}{\fontdimen8\scriptfont3}}}%
   \mkern-4mu\hbox{\let\f@size\sf@size\usefont{U}{lasy}{m}{n}\symbol{41}}}}
\makeatother

\def\eqref#1{equation~\ref{#1}}

\def\1{\bm{1}}

\def\m1{{\bm{1}}}

\DeclareMathAlphabet{\mathsfit}{\encodingdefault}{\sfdefault}{m}{sl}
\SetMathAlphabet{\mathsfit}{bold}{\encodingdefault}{\sfdefault}{bx}{n}



%% file: Dataset.tex
\vspace{-0.3em}
\subsection{Data Collection \& Annotation} \label{subsec:collection}




Building a dataset with open-ended questions and human-written descriptive answer is challenging because there are not many publicly available real-world sources with charts and related textual descriptions. After exhaustive search, we decided to use charts from Pew Research ({\href{https://www.pewresearch.org}{pewresearch.org}}). Pew serves as a suitable source because the articles are written by professional writers covering opinions, market surveys, demographic trends and social issues. The articles are often accompanied by a variety of real-world charts and their summaries. 

{We collected 9,285 chart-summary\change{-article triples} scraped from nearly 4,000 articles.} 
However, not all of the charts are suitable for creating open-ended questions. For example, some charts maybe too unconventional or too complex while a few others have poor resolution. Similarly, the text accompanying the chart may not discuss data values in the chart and instead refer to other external background facts. Hence, we manually went over all the charts to retain 7,724 samples that we deemed suitable for our study. In particular, we filtered out 1,019 samples as too complex and 542 as samples we cannot make an open-ended question. 

We perform an annotation study on the collected chart data to create question-answer pairs following the four steps below (see Table \ref{analysis-process} for  an illustrative example). More details of the data collection and annotation process are provided in \Cref{app:data-collect-annotate}.


\vspace{-0.5em}
\paragraph{(1) Question-answer Creation} We asked each crowdworker from Amazon Mechanical Turk to answer three existing 
questions (created by another crowdworker) \change{for three separate charts respectively,} and create three new question-answer pairs \change{for three new charts}. They were provided with the chart and the summary, and were asked to select {portions} of the text as an answer to the question. \change{The selected segments can be noncontiguous.} In this way, we collected two answers from different \change{workers} for each question, to verify answers and to remove any potential bias in answer selection.  


\vspace{-0.5em}
\paragraph{(2) Question Validation and Editing} After collecting the question-answer (QA) pairs, 
this and the next two steps are performed by five \change{internal} annotators who are native speakers of English \change{and have research background in summarization. Each QA pair is first examined by an  annotator who} 
first checks if the question is open-ended in nature, and edits the questions when the question is vague or incomplete, or not answerable from the charts.
Then, {the remaining annotators} analyze the questions in terms of grammatical correctness and edit them as needed. \change{Overall, the question was edited in 53\% question-answer pairs. In this 22.7\% cases were minor changes (less than 30\% tokens changed), 15.5\% cases were moderate changes (between 30\% and 60\% tokens changed) and 14.8\% were major changes (over 60\% tokens changed).}



\vspace{-0.7em}
\paragraph{(3) Disagreement Resolution}

As mentioned, we obtain two answers {from the crowdworkers} for each chart-question pair. To resolve any potential bias from one and/or disagreement between the two answers, we build an annotation interface where an annotator can either choose one of the two answers, or select a new answer from the given summary. The annotator checks whether the answer contains irrelevant information to the question or any text that are not derivable from the chart (\eg\ background information). For 18.4\% cases, the two answers matched exactly.  
For 68.2\% samples, the two answers still had high overlaps (over 90\% token matches); for another 10.1\% the overlaps between the answers were moderate (between 30\% and 90\% token matches) and for the remaining 3.3\%, the token matches between answers were less than 30\%. While resolving the disagreements between crowdworkers, in 96\% cases the annotators chose one of the two answers 
while for other 4\% they selected a new answer from the summary.



\vspace{-0.5em}
\paragraph{(4) Decontextualization}
In some cases, \change{crowdworkers} may have left out important information from the summary that is relevant to the question while in other cases, they may have included information that is not derivable from the chart. 
Thus, 
after 
selecting the most appropriate answer, annotators edit it {further} by adding \change{tokens from the summary} 
or removing tokens as necessary, \change{which is taken as the \textit{extractive answer} for the dataset}.
{Also, if needed, they replace the occurrence of a pronoun with its antecedent (a proper noun) in cases  where the entity is unknown, to put the answer in context, \change{which is the \textit{abstractive answer}}.}

\vspace{-0.5em}
\subsection{Dataset Analysis} 
\label{sec:dataset-analysis}



\begin{figure*}
     \centering
     \begin{subfigure}[b]{0.3\textwidth}
         \centering
         \resizebox{1.1\columnwidth}{!}
         {\begin{tabular}{lcc}
            \toprule
            \textbf{Statistics (on average)} \\
            \midrule
            Tokens in article & & 1268.91 \\
            Tokens in summary  & & 123.35 \\
            Tokens in title & & 17.94 \\
            Tokens in question & & 11.88 \\
            Tokens in abstractive answer & & 56.41 \\
            Tokens in extractive answer & & 56.21 \\
            Percentage of tokens extracted from the summary & & 52\% \\
            \change{Percentage of tokens extracted from the article} & & 7\% \\
            \midrule
            \textbf{Type} & \textbf{Simple} & \textbf{Complex} \\
            \midrule
            Bar &       712 &       4,823 \\
            Line &      234 &       1,667 \\
            Area &      7 &         4 \\
            Scatter &   0 &         42 \\
            Pie &       235 &       0 \\  
            \midrule
            Total &     1,188 &     6,536 \\
            \bottomrule
        \end{tabular}}
         \caption{\small Dataset Statistics and Chart Types}
        \label{dataset_stats_n_ctypes}
     \end{subfigure}
     \hfill
     \hfill
     \hfill
     \hfill 
     \begin{subfigure}[b]{0.6\textwidth}
         \centering
         \includegraphics[width=1\textwidth,trim={0cm 0.2cm 0cm 0cm}]{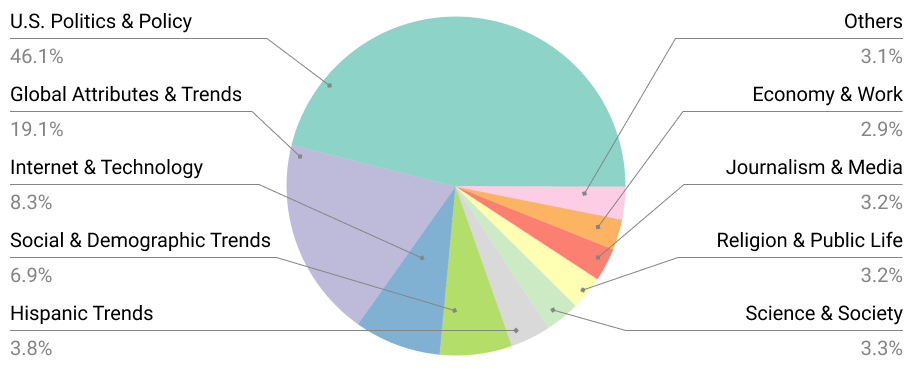}
         \caption{\small Distribution of topics.}
        \label{topics}
     \end{subfigure}
     \hfill
     \vspace{-2mm}
     \caption{\small (left) Dataset Statistics and Chart Types, (right) Topic Distribution}
    \label{dataset_stats_n_ctypes_n_topics}
    \vspace{-4mm}
\end{figure*}

\Cref{dataset_stats_n_ctypes} represents some basic statistics about the dataset. The questions and titles are generally short with both under 21 tokens on average. The percentage of tokens overlapping between the \change{extractive} answer and \change{the  article is 7\%  on average.}
Other characteristics of our dataset are as follows.






\vspace{-0.5em}
\paragraph{$\bullet$ 
Chart Types and Topics}
Our dataset contains a variety of chart types (\Cref{dataset_stats_n_ctypes}). The most common is bar charts (71.7\%), for both simple as well as stacked and group bar charts. The next most common type is line charts (24.6\%). Other types include area charts, scatter plots and pie charts. The dataset also covers a diverse range of topics including politics, technology, society and media (\Cref{topics}); about half of the charts cover \emph{U.S. Politics \& Policy} due to the nature of the dataset.



\begin{table}[t!]
\centering
\scalebox{0.7}
{\begin{tabular}{p{1.4cm}p{7.5cm}p{.4cm}}
\toprule
\textbf{Type}   & \textbf{Example} & \textbf{\%}\\
\midrule
Identify     &   What are the current thoughts on direct democracy? &       37\%\\
Summarize 

&   Explain the distribution of people who know a transgender person? &      
37\%\\
Compare      &    Compare Americans and Germans views about the world economic leader? &      20\%\\
Discover      &    How do Americans' see the coronavirus statistics? &      6\%
\\
\bottomrule
\end{tabular}}
\vspace{-0.5em}
\caption{\small Example and distribution of question types among 100 randomly selected questions. \change{The corresponding charts of these examples are shown in \Cref{question_type_examples}.}
}
\vspace{-5mm}
\label{dataset-question-types}
\end{table}

\vspace{-0.5em}
\paragraph{$\bullet$ Question Types}

We further analyze the question types using 100 randomly sampled question-answer pairs from our dataset.  \Cref{dataset-question-types} shows the distribution of questions across four main types. \change{Our categorization of questions are based on the specific analytical tasks with visualizations  one would have to perform to answer the question \cite{munzner2014visualization}. The four categories are: \Ni \textit{Identify:} 
questions that require identifying a specific target 
(\eg\ a data item or a data attribute) from the chart and describing the characteristics of that target; \Nii \textit{Compare:} questions that require comparisons between specified targets from the chart;
\Niii \textit{Summarize:} questions  that require summarizing the chart based on specified statistical analysis tasks (\eg\ describing data distribution,  outliers(s), and trend(s)); 
and \Niv \textit{Discover:} questions that require analyzing the whole chart to derive key insights through inference and reasoning. Unlike the \textit{summarize} task there is no explicit analytical task that is specified in a \textit{discover} type question.}

From \Cref{dataset-question-types}, we notice that people often 
ask to 
locate one or more portions of a chart (identify and compare)  and then characterize them to answer the question. They may also ask 
for a descriptive answer of some trends or data distributions. In contrast, questions that require the user to focus on the whole chart (\eg\ discovery) are fewer. This suggests that unlike the chart 
summarization problem which focuses on 
the whole chart, OpenCQA problem requires the model to identify relevant portions of the chart to answer the question.

%% file: Models.tex
\vspace{-0.3em}
\paragraph{Problem Definition}

\noindent For our OpenCQA problem, we consider \change{three} task settings. \change{In the first setup, the model takes a chart and the article containing the chart as input and extracts an answer to an open-ended question. The data for this task can be represented as a tuple of 6 elements $\mathcal{D} = \{\langle C,T,M,Q,D,A\rangle_n\}_{n=1}^{N}$, where $C,T,M,Q,D$ and $A$ represent the chart image, title, metadata, question, document (article) text and answer text, respectively.  The metadata $M = \langle C_{\text{label}},C_{\text{bbox}} \rangle$ consists of the chart labels \change{that are text segments extracted from the chart through OCR (\eg\ axis labels, data labels)} and their respective bounding boxes. In the second setup, the chart summary is provided as input instead of the whole article.} The dataset in this setup can be represented as $\mathcal{D} = \{\langle C,T,M,Q,S,A\rangle_n\}_{n=1}^{N}$, where $S$ represents the chart summary. In the \change{third} setup, the chart summary is not accessible, and the model must rely only on the chart. This is a  more difficult and interesting problem setting since real world charts often do not come with explanatory summaries. {In this setting, for an input $I = \langle C,T,M,Q \rangle$, the model has to learn to generate an explanatory answer $A$.} 


For training the models, we use state-of-the-art \emph{extractive} and \emph{generative} QA models. 
{The supervision of extractive models are \emph{extractive} answers 
whereas for the generative models \emph{abstractive} or  edited (as described in \Cref{subsec:collection}) answers are used.}
Note that the \change{third} task setting applies to only generative models, whereas the \change{first and second} settings apply to both extractive and generative models. 
We describe the models below (implementation details are in \Cref{app:baselines}).

\subsection{Extractive Models}

We adopt two extractive models for the \change{two} problem setups where the models extract the answer to the question from the input 
summary \change{or article}.

\noindent \textbf{$\bullet$ BERTQA}
\cite{chadha2019bertqa}
is an extractive QA model that 
uses 
directed coattention layers~\cite{xiong2016dynamic} 
to improve the performance of the original BERT model \cite{devlin-etal-2019-bert}. In this approach, we first pass the question and the text (\change{article or summary})
through a BERT model to get the (self-attention based) representations. It then calculates the cross attention from the question to \change{text and text} to question and concatenates the resulting vectors to predict the start and end points of the answer span. 


\noindent \textbf{$\bullet$ ELECTRA}
\cite{clark2020pretraining} proposes a
self-supervised representation learning method that emphasizes computational efficiency. In contrast to Masked Language Modeling (MLM), it uses {Replaced Token Detection} or RTD as the pretraining task. The training process is inspired by the training of Generative Adversarial Networks or GANs \cite{goodfellow2020generative}. The training examples are first passed to a generator (typically a small MLM-based model) to replace some of the tokens in the input with other probable but incorrect tokens. ELECTRA {(the discriminator)} is then trained to distinguish between the ``original'' vs. ``replaced'' tokens in the input. This binary classification task is applied to every token, hence it requires fewer training examples compared to MLM training. ELECTRA  achieves state-of-the-art results on SQuAD 2.0 \cite{rajpurkar2018know}. We use the same experimental setup as with the BERTQA model.



\subsection{Generative Models}
   
   
\begin{table*}[t!]
    \centering
    \scalebox{0.74}
    {\begin{tabular}{
     p{2cm}p{1.7cm}p{4cm}p{1.6cm}p{1.9cm}p{1.7cm}p{2.5cm}p{1.5cm}
    }
        \toprule
        \textbf{Models} & \textbf{Type} & \change{\textbf{ROUGE\_-F (1/2/L)}} $\uparrow{}$  & \textbf{CS} $\uparrow{}$ & \textbf{BLEURT} $\uparrow{}$ & \textbf{CIDEr} $\uparrow{}$ & \change{\textbf{BERTScore}} $\uparrow{}$ & \textbf{BLEU} $\uparrow{}$ \\[-.6em]
       \midrule
        \multicolumn{8}{c}{\textbf{\change{Setup 1: With Article Provided}}}  \\
\rowcolor{gray}        BERTQA & Extractive & 22.81/8.53/16.63 & 28.17\% & -0.692 & 0.983 & 85.00 & 9.58 \\
\rowcolor{gray}        ELECTRA & Extractive & 57.67/50.09/53.89 &  54.73 \% &  0.066 &   5.100  & 91.01 & 38.79 \\ 
        BART & Generative & 50.68/38.95/45.55 & 54.29\% & -0.017 & 4.121 & 91.08 &  23.91 \\
        T5 & Generative & 66.57/60.40/\textbf{63.46} & \textbf{68.24\%} & \textbf{0.103} & \textbf{5.437} & \textbf{93.53} & \textbf{41.28} \\
         \change{VLT5} & Generative & 58.90/49.97/54.82 & 66.06\% & 0.076 & 5.227 & 92.34 & 40.06 \\
        GPT2 & Generative & 16.71/66.57/13.75 & 14.71\% & -0.745 & 0.814 & 82.05 & 0.99 \\
        CODR & Generative & 14.58/1.31/11.37 & 3.4\% & -1.155 & 0.051 & 81.90 & 0.43 \\
        \midrule
        \multicolumn{8}{c}{\textbf{Setup 2: With Summary Provided}}  \\
\rowcolor{gray}        BERTQA & Extractive & 70.97/67.77/69.34 & 85.28\% & 0.297 & 7.177 & 94.65 & \textbf{66.33} \\
\rowcolor{gray}        ELECTRA & Extractive & 76.24/74.33/\textbf{75.48} & \textbf{92.57\%} & \textbf{0.378} &  \textbf{7.823} & \textbf{96.06} &  65.40 \\ 
        BART & Generative & 66.41/61.86/64.41 & 68.71\% & 0.257 & 6.613 & 93.85 &  38.42 \\
        T5 & Generative & 75.77/73.51/74.51 & 81.73\% & 0.369 & 7.728 & 95.22 & 57.93 \\
        \change{VLT5} & Generative & 75.07/72.35/73.86 & 85.36\% & 0.376 & 7.597 & 95.20 & 59.80 \\
        GPT2 & Generative & 14.29/11.13/13.42 & 22.00\% & -0.782 & 1.725 & 49.18 & 12.68 \\
        CODR & Generative & 13.81/0.76/10.25 & 2.480\% & -1.039 & 0.038 & 81.87 & 0.31 \\
        \midrule
        \multicolumn{8}{c}{\textbf{Setup 3: Without Summary Provided}} \\
        BART & Generative & 40.29/21.40/32.48 & 49.07\% & -0.166 & 2.260 & \textbf{89.69} & 7.41 \\
        T5 & Generative & 41.12/22.09/32.97 & 52.30\% & -0.173 & 2.357 & 89.59 & 9.28\\
        \change{VLT5} & Generative & 42.87/22.60/\textbf{33.29} & \textbf{54.47\%} & \textbf{-0.134} & \textbf{2.447} & 89.53 & \textbf{14.73} \\
        GPT2 & Generative & 28.55/11.26/22.46 & 32.00\% & -0.493 & 1.314 & 85.05 & 4.89 \\
       CODR & Generative & 14.67/1.05/10.90 & 4.14\% & -1.170 & 0.053 & 81.86 & 0.32 \\
        \bottomrule 
    \end{tabular}}
    \vspace{-2mm}
    \caption{\small Evaluation results for different models on the test set. $\uparrow{}$ means higher is better, $\downarrow{}$ means lower is better. {All models use the OCR-extracted data and chart title in the input.
    70:15:15 split is used for train:test:val. 
    }  
    \vspace{-4mm}
    }
\label{tab:evaluation-table}
\end{table*}

\textbf{$\bullet$ GPT-2} \cite{radford2019language} 
trains a transformer decoder on the unlabelled BooksCorpus dataset \cite{zhu2015aligning} using a conditional language modelling objective. Their pretrained model can be fine-tuned on downstream tasks such as textual entailment, similarity and question answering. 
We fine-tune GPT-2 under \change{the three} tasks where {all the input elements in each task setting} are concatenated as conditioning input to predict the answer. 


\noindent  \textbf{$\bullet$ BART}
\cite{lewis-etal-2020-bart} uses a standard encoder-decoder transformer architecture. Its pretraining task involves denoising, where text spans in the input text are replaced with a single mask token, and the decoder is tasked to predict the original input sequence. BART has been shown to achieve state-of-the-art performance on text generation tasks such as summarization.
\change{In each of our three task settings, we concatenate the corresponding inputs and feed into the model for fine-tuning.}


\noindent  \textbf{$\bullet$ T5}
\cite{Raffel2020t5} is a unified encoder-decoder transformer 
model that converts language processing tasks into a text2text generation format. It is first pretrained with a `fill-in-the-blank' denoising objective, where 15\% of the input tokens are randomly dropped out. The spans of consecutive dropped-out tokens and dropped-out tokens that stand alone are then replaced by special sentinel tokens. 
Each sentinel token is assigned a token ID that is unique in the input sequence. The  decoder then learns to predict those dropped-out tokens, delimited by the same input sentinel token plus a final sentinel token. We fine-tuned T5 on our tasks using the same input format as with BART.


\noindent  \textbf{$\bullet$ VLT5}
\cite{cho2021unifying} \change{is a T5-based framework that unifies the Vision-Language (VL) tasks as text generation conditioned on multimodal inputs. The input consists of both textual tokens and visual features of objects from the image extracted by Faster R-CNN \cite {renNIPS15fasterrcnn}. The model is pre-trained on multiple multimodal tasks: language modeling, visual QA, visual grounding, image-text matching and grounded captioning. We fine-tuned VL-T5 on our OpenCQA generative task in the following manner. For the textual input, we use the same input format as T5. For the visual input, we extract the visual features of different marks in the chart image (\eg\ bars, lines) using Mask R-CNN \cite {8237584} with Resnet-101 as its backbone.}


\noindent \textbf{$\bullet$ CODR}
\cite{prabhumoye-etal-2021-focused} proposes a document grounded generation task, where the model 
uses the information provided in a document to enhance text generation. In their setup, the context and source documents are concatenated and passed to a BART encoder to get a contextualized representation of the document. Then the same encoder is applied to the context alone and both representations are finally concatenated and passed to the BART decoder to generate the text. \change{We fine-tune {the pretrained BART encoder and decoder following the approach of CODR} for OpenCQA
by taking the question as the context
and concatenating the rest of input elements as the grounded document.}

%% file: Evaluation.tex

 \vspace{-0.2em}
\subsection{Automatic Evaluation}

\paragraph{Measures} We utilized six measures for automatic evaluation. 
{\textbf{BLEU} \cite{post-2018-call} measures $n$-gram overlaps between the model generated text and the reference. It is a precision oriented measure and is 
calculated by taking the geometric mean of BLEU1-4 and multiplying by the exponential of brevity penalty. \change{\textbf{ROUGE} \cite{lin2004rouge}, on the other hand, is recall based and it  
measures $n$-gram overlaps between the model generated text and the reference as a percentage of $n$-grams in reference summaries. We report ROUGE-F score.}

\textbf{CIDEr} \cite{vedantam2015cider} measures TF-IDF weighted $n$-gram overlaps between the model generated text and the reference.}
\textbf{BLEURT} \cite{sellam-etal-2020-bleurt} is a model-based 
metric that measures the fluency of the generated sentence and the extent to which it conveys the meaning of the reference. We use BLEURT-base-128. 
\textbf{Content Selection or CS} score \cite{wiseman-etal-2017-challenges} measures how well the generated text matches the 
gold answer 
in terms of selecting which records to generate. 
Since both the BLEURT and CS are calculated at the sentence-level,  
we 
average these scores over the whole test set.
\change{Finally, we use \textbf{BERTScore} \cite{zhang2019bertscore} which correlates better with human judgments and has been widely adopted recently for text generation evaluation.  BERTScore computes token similarity  between the candidate and reference using contextualized embeddings rather than relying on exact matches.} 

\begin{figure*}[t!]
\linespread{0.5}\selectfont\centering 
\renewcommand{\arraystretch}{1.0}
 \scalebox{0.84}
{\begin{tabular}{p{4.9cm} | p{5.3cm} | p{5.7cm}}     
        \small \underline{\textbf{Q1: }} Describe the change in favorability towards Congress.
        \raisebox{-1.05\height}{\hspace{-17mm} \includegraphics[height=4.5cm, trim={10.5cm 5.7cm 6cm 6.5cm}, clip]{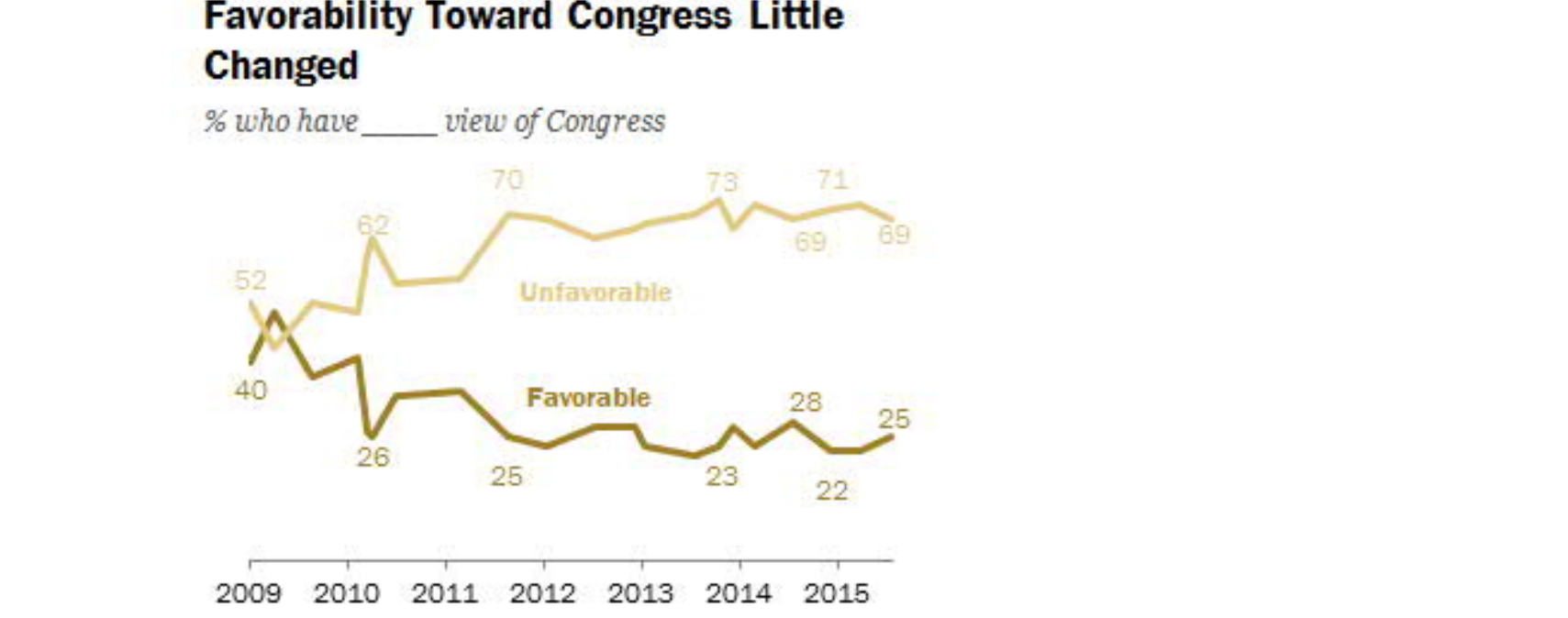}}
        &
       \small \underline{\textbf{Q2: }}
         Describe the proportions of U.S. adults and AAAS scientists that think population growth won't be a problem.
        
        \raisebox{-1\height}{ \hspace{-3mm} \includegraphics[height=3.9cm, trim={6.9cm 5cm 3cm 1cm}, clip]{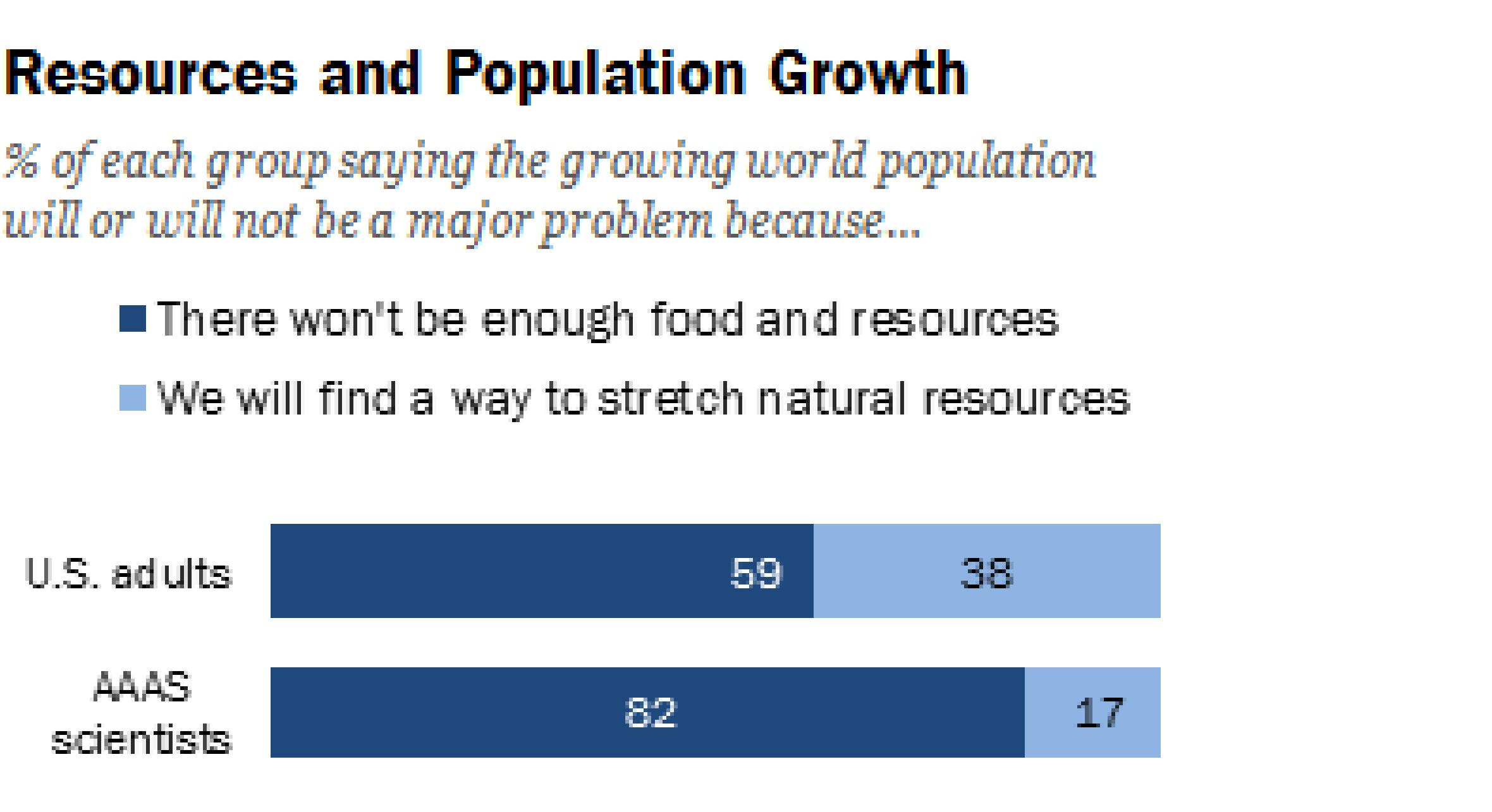}}
        &
       \small \underline{\textbf{Q3: }}
        How is the relation between family income and early technology adoption?
        
        \raisebox{-1.0\height}{ \hspace{0mm} \includegraphics[height=4.8cm, trim={9.5cm 0cm 7cm 1.5cm}, clip]{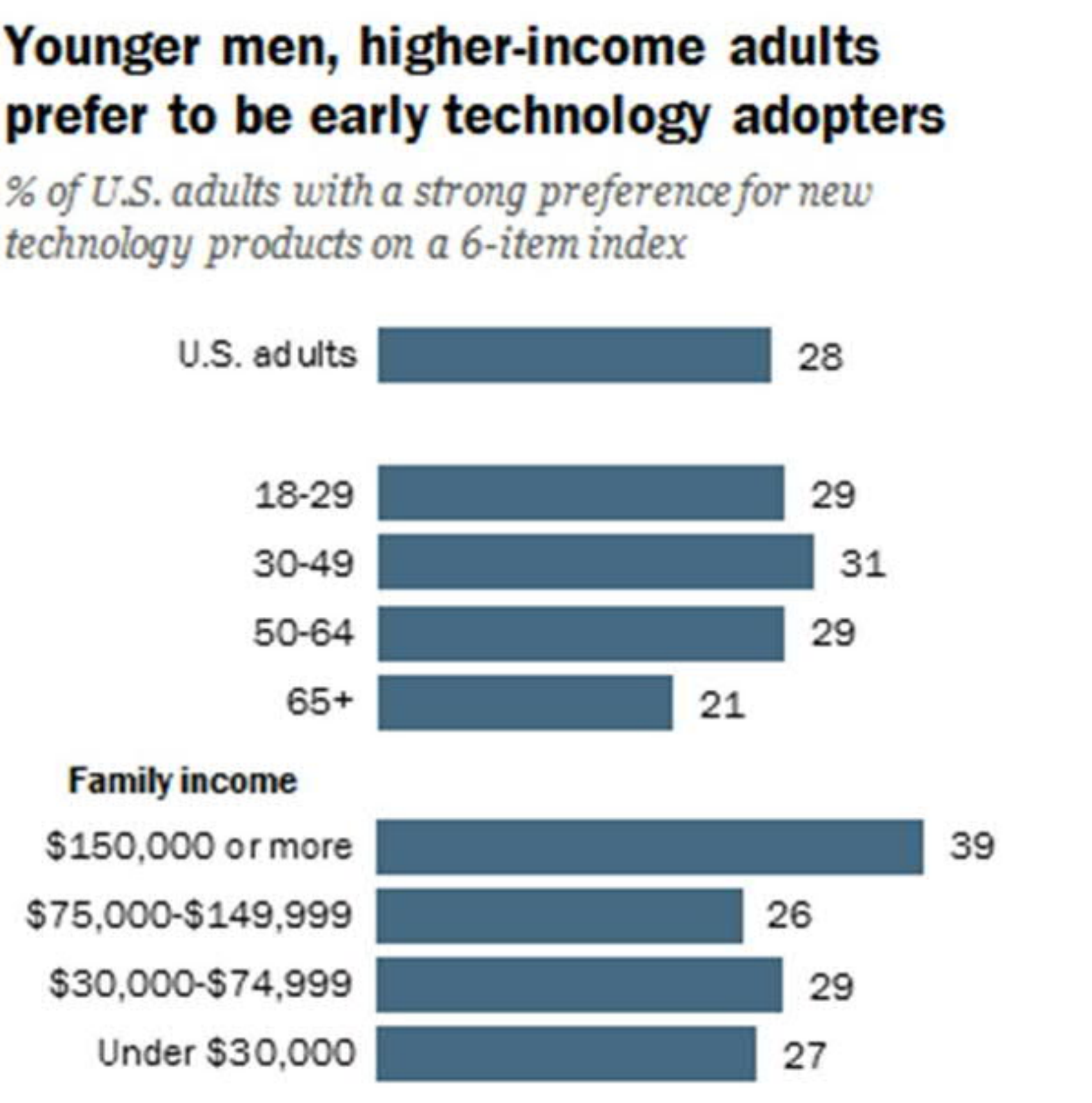}}
        \\
\midrule

        {\small \textbf{Gold:} Views of Congress continue to hover near record lows: Just 25\% of Americans view Congress favorably. }
        &
        {\small \textbf{Gold:} Just 17\% of AAAS scientists and 38\% of Americans said population growth won’t be a problem because we will find a way to stretch natural resources.}
        &
        {\small \textbf{Gold:} Adults with family incomes of at least \$150,000 annually  tend to express a stronger preference for early technology adoption on this summary index (39\% do so) compared with those at middle and lower income levels}
        \\
\midrule 

        
    
    {\small \change{\textbf{VLT5:}} 25\% express a favorable opinion of Congress, while 69\% view it unfavorably. But since April 2014, the share of Americans with a positive view of Congress has fallen off to 25\%.
}
        & 
        
        
        {\small \change{\textbf{VLT5:}} Fully \red{59\%} of AAAS scientists say population growth will n’t be a major problem because there will not be enough food and resources in the world, and 17\% say we will find a way to stretch natural resources.}
        &
        {\small \change{\textbf{VLT5:}} Younger men (\red{39\%}) are more likely than older adults (\red{26\%}) to express a strong preference for being early technology adopters.
}
        \\
\midrule 

        
         {\small \change{\textbf{Bboxes-VLT5:}} A majority of the public (\red{69\%}) has a favorable opinion of Congress, while \red{23\%} view it unfavorably. In July 2014, the balance of opinion was just the opposite: \red{47\%} expressed a favorable opinion of Congress, while \red{25\%} held an unfavorable view.}
        &
        
        {\small \change{\textbf{Bboxes-VLT5:}} \red{Two-thirds} of AAAS scientists say population growth will n’t be a major problem because of scarce resources and other resources. At the same time, about \red{six-in-ten} of both groups say that the U.S. will find a way to stretch natural resources and hope that we will find a way to stretch these resources}
        &
        {\small \change{\textbf{Bboxes-VLT5:}} Older people in the U.S. tend to have higher levels of education and skills, rather than being early adopters. Many of those with higher incomes prefer early technology adoption; in fact, lower-income adults lag behind younger men in this regard.}
        \\
        

\bottomrule  
    \end{tabular}
    }
    \vspace{-2mm}
    \caption{\small Sample outputs from the Pew test set on VLT5 model (without summary provided).\change{"Bboxes-" models use bounding box information of chart elements.} \red{Red} indicates factual errors.
    }
    \label{fig:vlt5-outputs} 
    \vspace{-4mm}
\end{figure*}

\vspace{-0.5em}
\paragraph{Results}
\change{
As expected, the models perform better when only the relevant chart summary is given compared to when the entire article is provided (the first 2 setups in ~\Cref{tab:evaluation-table}). The generative models have reduced performance due to the added texts
in the article in setup 1.
The extractive models 
also have a loss in performance in this setup
mainly because they assume the answer comes from a single contiguous text segment whereas \change{a gold answer can be constructed from multiple segments (\Cref{subsec:collection})}. Specifically,  these methods apply a sliding window over the tokens in the input sequence to find the start and end tokens of the answer span. However, if there are irrelevant tokens in the span, a window may miss portions of the answer that fall outside the maximum sequence length. Also, during training, the model rejects those instances where the answer span is not contained wholly inside the input sequence. The problem is worse with BERTQA because it uses a sequence length of 512 which is much shorter than the article size (1268 tokens on average), {whereas ELECTRA uses a sequence length of 2048 tokens, reducing this effect.}

}

 We notice that the models perform better when the summary is provided compared to when only the chart is provided (setup 2 vs. 3 in ~\Cref{tab:evaluation-table}). This is expected because when the summary is not given, the model does not have any textual reference to generate the answer to a given question. The extractive models BERTQA and ELECTRA achieve high \change{ROUGE-L}  since they directly extract the text from the given summary to answer the question. \change{ELECTRA performs better than BERTQA on all metrics except BLEU {showing the benefit of the generator-discriminator based cascaded training.}} 

When the summary is not provided, \change{VLT5} 
achieves the best results across most measures, likely because it uses both visual and textual features to generate answers.
On the contrary, CODR performs poorly possibly because it was originally intended for discourse generation whereas chart QA is quite a different task. GPT-2 performs better than CODR but it has lower \change{ROUGE-L} and CS scores compared to BART and T5.

\change{We conduct several additional experiments to understand: \Ni how performance may vary with respect to different chart types; \Nii how including the bounding box information as input impacts the performance; (iii) how naive baselines that either randomly select the answer from the input text or only take the question/summary as input perform; and \Niv how much the OCR-extracted text contributes to the output. Details
of these experiments can be found in \Cref{app:additional}.} 

\begin{table}[t!]
\vspace{-0.3em}
\setlength{\tabcolsep}{3pt}
    \centering
    \small
    \scalebox{0.75}{\begin{tabular}{lccccc}
    \toprule
   
    & \textbf{Factual} & \textbf{Relevance} & \textbf{Fluency} & \textbf{\change{Content Cov.}} & \change{\textbf{CS}}\\
   
    \midrule
    \change{VLT5} &    44.89\% &    86.67\% & 92.45\% &    41.11\% & 53.19\%\\
    \change{VLT5-S} &    82.45\% &    88.67\% &   98.45\% & 85.12\% & 81.15\%\\
    Gold &           96.45\% &   96.00\% &   98.89\% &    100.00\%  & 100.00\%\\
    \midrule
    Krippendorff's $\alpha$ & 0.75 & 0.76 & 0.83  & 0.79 &\\
    \bottomrule
    \end{tabular}
    }
    \vspace{-2mm}
    \caption{\small Human evaluation results for comparing between the outputs of VLT5 (without summary), VLT5-S (with summary) and the gold answer. The last row shows the agreements.%
    }
    \label{tab:evaluation-table-human}
    \vspace{-5mm}
\end{table}

\vspace{-0.5em}
\subsection{Human Evaluation}
To further assess the quality of the answers generated by the models, we performed a human study on 150 randomly sampled charts from our OpenCQA dataset with three internal annotators who are native speakers of English. For comparison, we chose \change{the outputs from} the best performing generative model (VLT5) \change{on the with and without summary settings, which we denote as VLT5-S and VLT5, respectively.} 
We also provide the gold answer (as control)  and randomise the order of all three answers to remove biases. The annotators assess each \change{answer} based on three criteria adapted from \cite{yuan2021bartscore}: \Ni \textbf{Factual correctness} which measures the extent to which the generated text contains statements entailed by the source (\ie\ generated facts are supported by the chart), \Nii \textbf{Relevance} which measures how consistent the generated text is with respect to the question, \Niii \textbf{Fluency} which measures the extent to which the generated text is deficient of formatting issues, capitalization errors or ungrammatical sentences (\eg\ fragments, missing components) which would otherwise make the text difficult to read. For these three criteria, annotators provided a 5-point Likert scale rating from 1 (the worst) to 5 (the best).

\change{We further propose a new \emph{recall-oriented} criteria called \textbf{Content Coverage} which measures the extent to which the facts from the gold answer are covered by the generated text, ranging from 1 (0-20\%) to 5 (80-100\%). For this, the order of outputs of the two models were randomized to counterbalance any potential ordering effect.}
Each comparison was performed by three annotators. To measure inter-annotator reliability, we computed  Krippendorff’s alpha \cite{krippendorff2011computing} and found a good overall agreement with  an alpha 
coefficient of \change{0.806.}

Table \ref{tab:evaluation-table-human} shows the results of human evaluation, where for each model we show the percentage of \change{answers} that received the nearly perfect/perfect ratings (\textit{i.e.,} 4 and 5). 
The \change{gold answers} receive high ratings across \change{the first} three measures which attests the quality of our dataset. 
\change{VLT5-S achieves higher ratings across all measures than VLT5 
as it has access to the summary. 
The VLT5 model also received impressive ratings on Relevance and Fluency while it has more room for improvement on Factual Correctness and Content Coverage, especially when the summary is not given. Lastly, we see that the automatic and the human evaluation measures for content coverage, \ie\ the CS score and Content Coverage, both of which are recall oriented, tend to be correlated.}

\vspace{-0.5em}
\subsection{Error Analysis and Challenges}
\label{sec:error-analysis}

We have manually analyzed the 
outputs 
from our best performing model to explain the key challenges that existing models face. Overall, the 
outputs of \change{VLT5} model are often fluent and grammatically correct as evident from \Cref{fig:vlt5-outputs}. Also, this model rarely hallucinates and is often true to the source context.
However, there are still several areas for improvement which we  discuss below.
\\
\noindent \textbf{Factual Correctness and relevancy} In \Cref{fig:vlt5-outputs},   both variants of \change{VLT5} fail to associate some reported data values (highlighted in red) with their corresponding axis labels. In (Q3 in \Cref{fig:vlt5-outputs}), the \change{VLT5 model outputs} some facts about the relation between `age' and `early technology adoption' but the question asked about the relation with `family income', not `age'. To address the factual errors and irrelevant facts, future models could create better input representations including semantic graph representations \cite{Teney2017GraphStructuredRF} by extracting and exploiting  relationships among the question, data values, chart elements (\eg\ axis labels) and their visual features (\eg\ colors).
\\
\noindent \textbf{Mathematical \& Logical Reasoning} Often gold answers include numerical reasoning to provide rich explanations. For example,  in~\Cref{fig:vlt5-outputs}(Q2), the gold answer provides a sense of limited proportion by mentioning \change{``Just 17\% of AAAS...''}, which is not captured in the generated outputs. Future models may improve such numerical reasoning as an intermediate step  before surface realization \cite{chen2020logic2text}. In another case (Q3 in~\Cref{fig:vlt5-outputs}), the gold answer considers that incomes below 150k are middle and lower income levels which is not captured by models. Future work could infer such rich semantics with the help of external knowledge sources and inference based modelling \cite{bowman-etal-2015-large,talmor-etal-2019-commonsenseqa}.\\
\noindent  \textbf{Rare Topics} We also observe that the \change{VLT5} model tends to produce less fluent and factually correct answers when the sample comes from a rare topic (\eg\ \change{Bboxes-VLT5 in \change{Q2}} in ~\Cref{fig:vlt5-outputs}). Future work could focus on transfer learning and building more diverse datasets to handle such variations. \\
\noindent \textbf{Automatic Evaluation} It has been acknowledged that metrics like BLEU do not capture many data-to-text generation issues~\cite{parikh2020totto}. This is especially the case for open-ended questions where correct answers can be expressed in multiple possible ways.
While we evaluated the models through several metrics, we urge for further research on building better automatic metrics that capture model performance in terms of factual correctness, relevancy and reasoning aspects \change{while also correlating strongly with human judgements}.

%% file: Appendix.tex
\appendix

\section{Appendices}
\begin{figure*}[t!]
    \begin{subfigure}[t]{.3\textwidth}
         \includegraphics[height=1.3in,trim={0cm -1cm 0cm 0cm},clip]{ 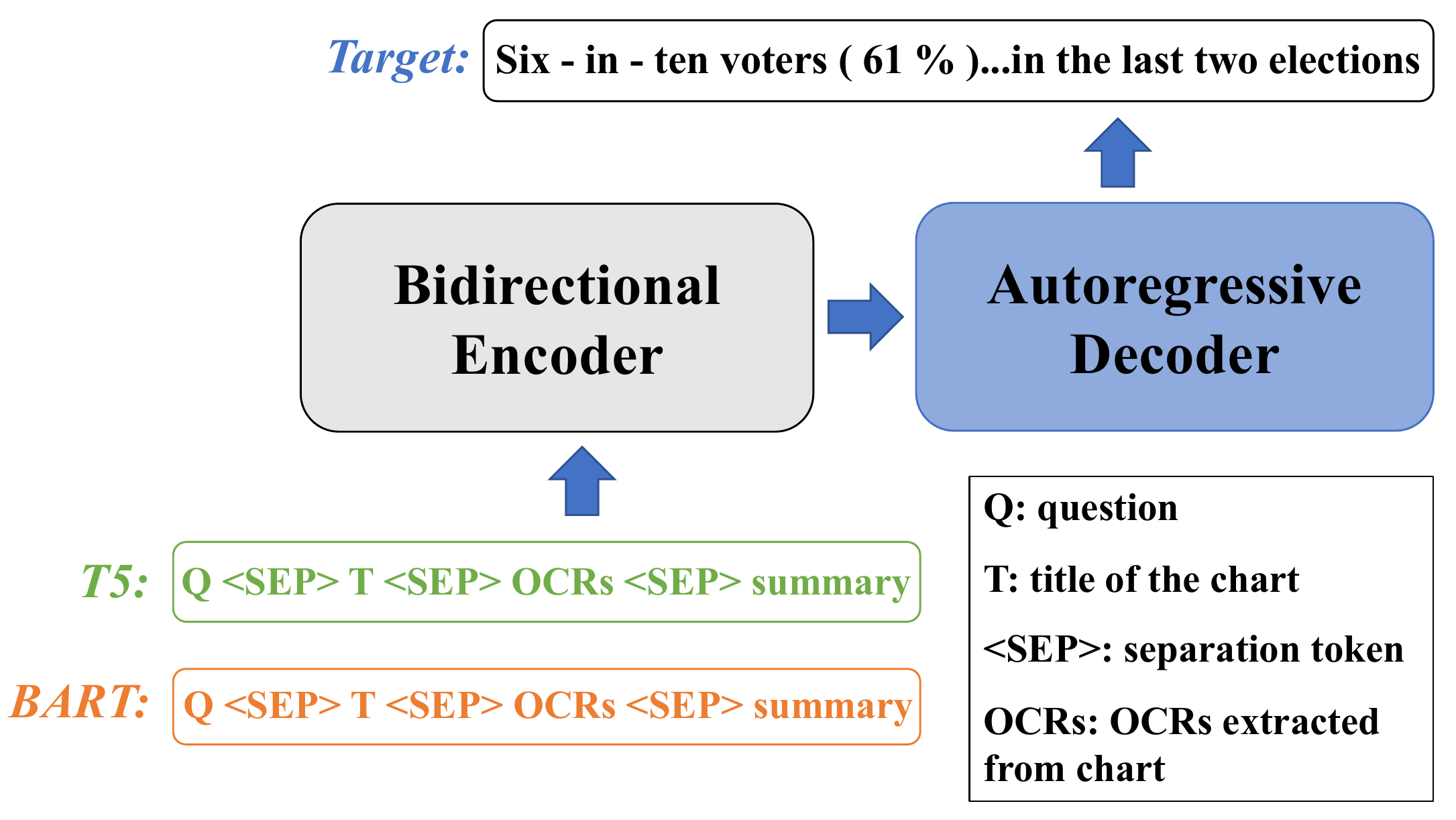}
        
         \caption{\small T5 and BART model. }
        \label{t5_sample}
     \end{subfigure}
     \hfill
    \begin{subfigure}[t]{.3\textwidth}
         \includegraphics[height=2in,trim={15cm 0cm 2cm 11cm},clip]{ 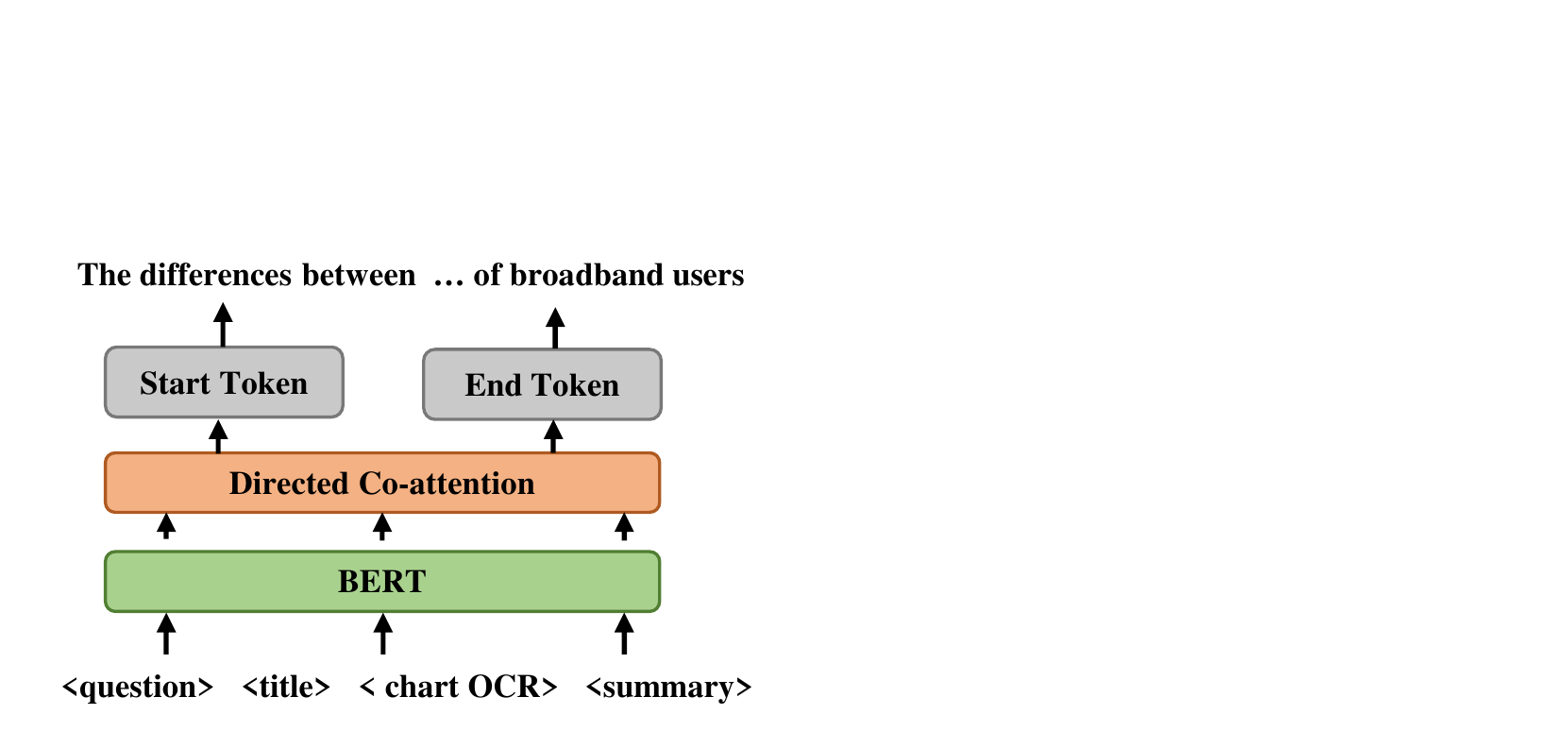}
        
         \caption{\small BERTQA model.}
        \label{bertqa_sample}
     \end{subfigure}
     \hfill
     \begin{subfigure}[t]{.3\textwidth}
         \includegraphics[height=2in,trim={14cm 0cm 3cm 11cm},clip]{ 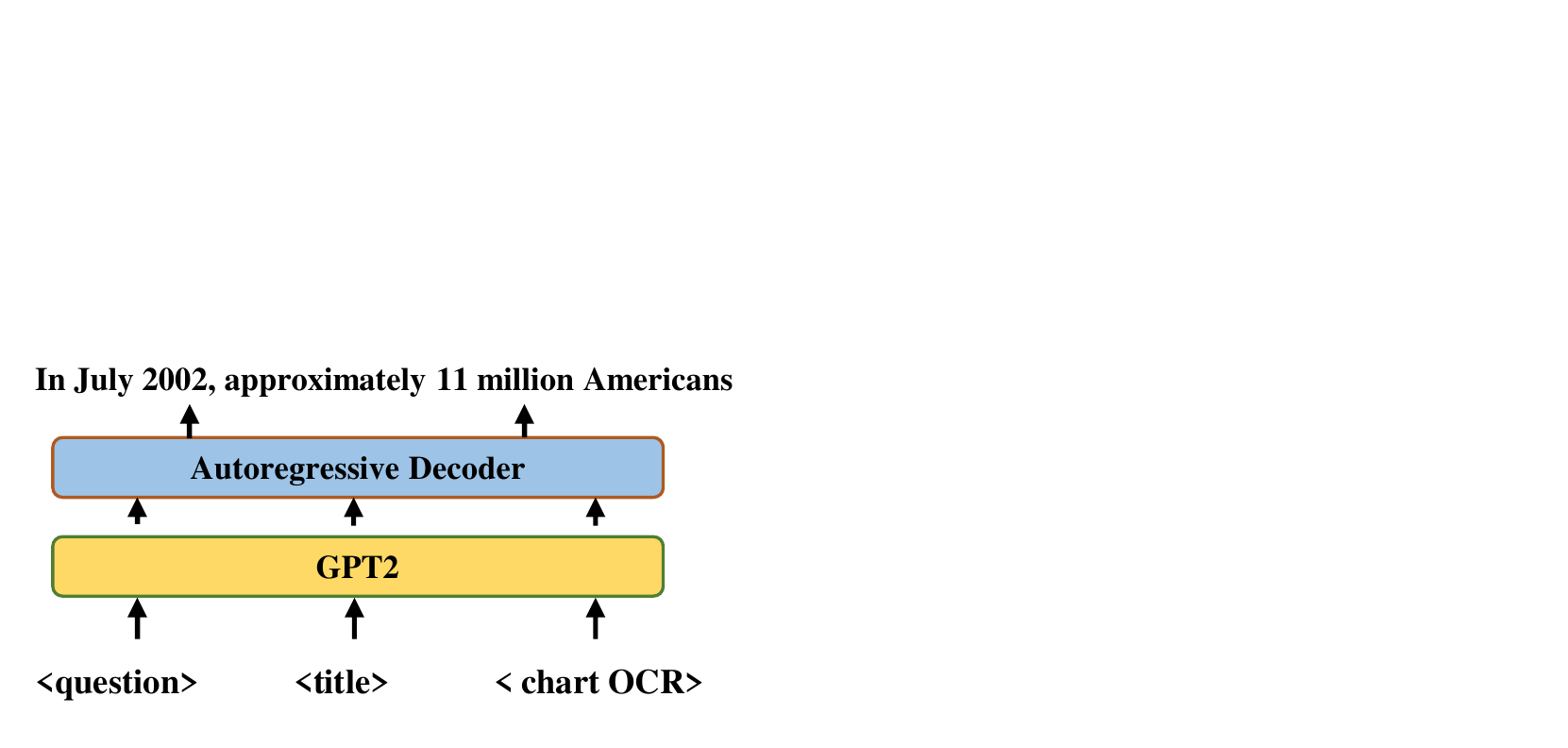}
        
         \caption{\small GPT2 model 
         }
        \label{gpt2_sample}
     \end{subfigure}
     \caption{\small Examples of model architectures that show the fine-tuning step of the training stage with sample inputs and outputs. (a) and (b) show examples of inputs and outputs from models when the summary is given.  (c) show examples of inputs and outputs of GPT2 (when the summary is not given as input). Here, BERTQA is an extractive model, others are generative models.
     }
    \label{sample_models}
\end{figure*} 

\subsection{Data Collection \& Annotation}
\label{app:data-collect-annotate}

\subsubsection{Data Collection}  
\label{app:data-collect}

\begin{figure*}[b!]
\linespread{0.5}\selectfont\centering 
\renewcommand{\arraystretch}{1.0}
 \scalebox{0.8}
{\begin{tabular}{p{6.5cm} | p{6.8cm} | p{5cm}}     
\toprule
        \begin{center} \vspace{-4mm} \textbf{(1) Too Complex} \end{center}
        \raisebox{-1\height}{\hspace{-19mm}  \includegraphics[height=4.5cm, trim={10cm 2.3cm 0cm 10.1cm}, clip]{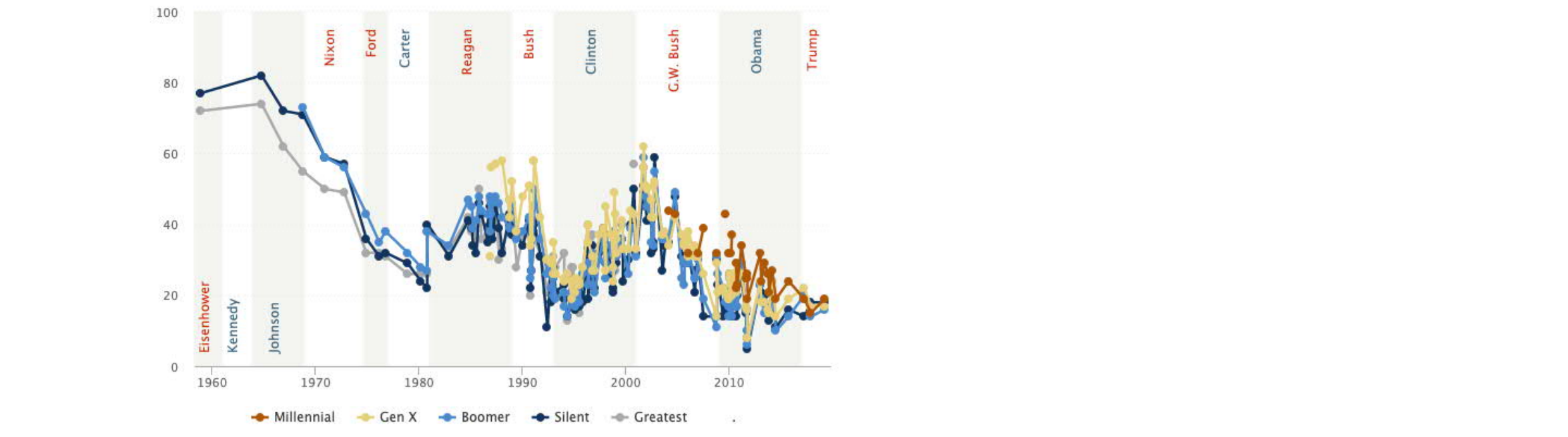}}
        &
       
        \begin{center}  \vspace{-4mm} \textbf{(2) Cannot Create Open Ended Question} \end{center}
        \raisebox{-1\height}{ \hspace{-3mm}  \includegraphics[height=4.5cm, trim={13.5cm 6cm 10cm 7cm}, clip]{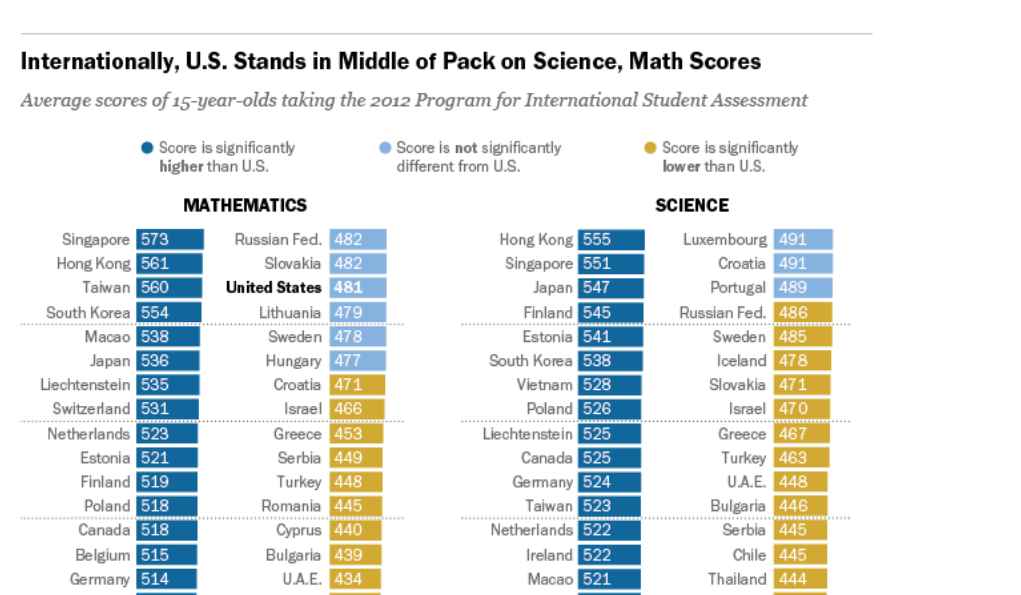}}
        &
       
         \begin{center} \vspace{-4mm} \textbf{(3) Good} \end{center}
         
        \raisebox{-1.0\height}{ \hspace{0mm} \includegraphics[height=5cm, trim={14.3cm 3cm 0cm 8cm}, clip]{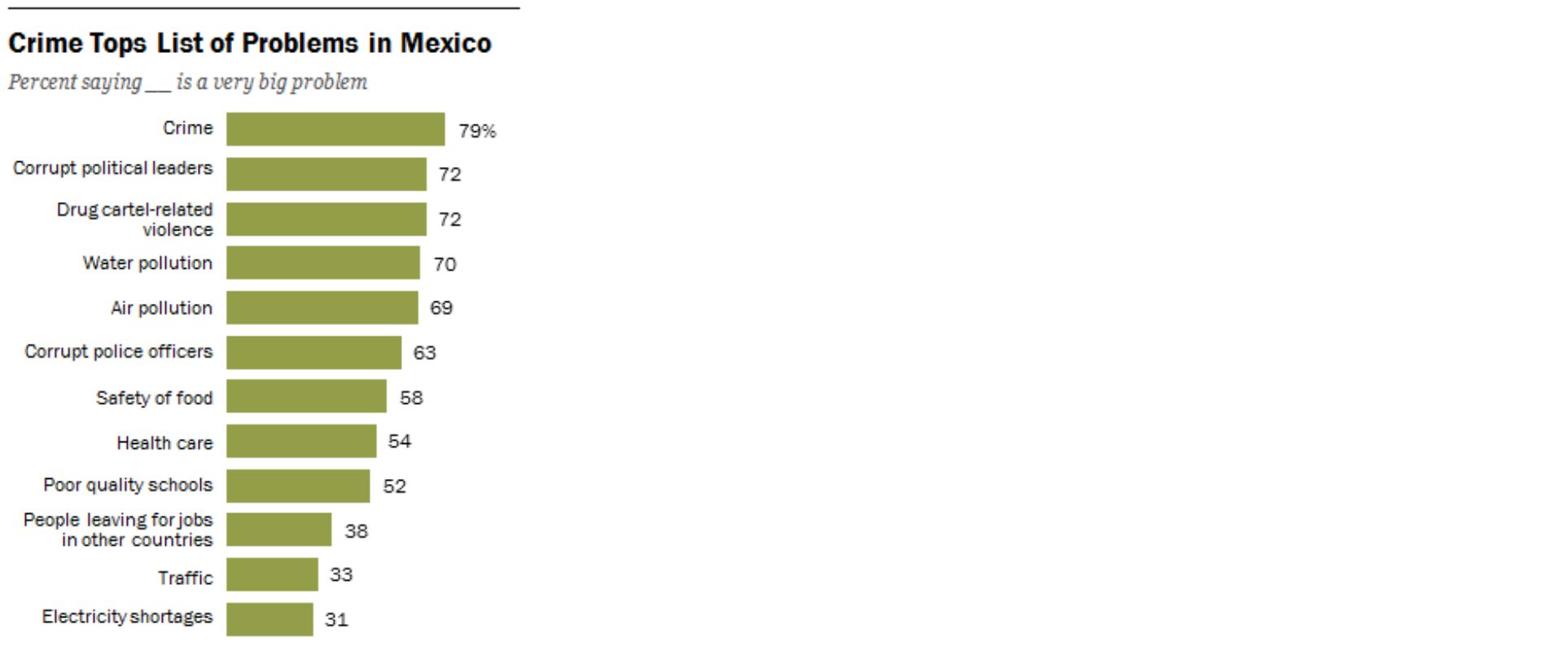}}
        \\

\midrule 

         {\small Historically, there have been modest differences between generational groups in trust in government and that remains the case today. Currently, 19\% of Millennials (now ages 23-38) report trusting the government, similar to the shares of older generations who say the same. Trust in government remains at or near historically low levels across generational lines.}
        &
        {\small Fact Tank’s look at standardized test results around the world seemed to bear out these sentiments: U.S. students are improving – slowly – in math and science, but still lagging their peers internationally.}
        &
        {\small About six-in-ten (58\%) say food safety is a very big problem, and 54\% say the same about health care. Roughly four-in-ten or fewer are troubled by people leaving for jobs in other countries (38\%), traffic (33\%) and electricity shortages (31\%).}
        \\
        

\bottomrule  
    \end{tabular}
    }
    \vspace{-0.5em}
    \caption{\small  
    Examples of charts in the Pew dataset: (1) too complex since the line charts are difficult to separate; (2) not suitable for creating an open-ended question since summary does not refer to information in the chart; and (3) a good sample consisting of a chart and  the summary that refers to the chart information. We filter out samples like (1) and (2) from our dataset.
    }
    \label{chart_filtering_examples} 
\end{figure*}

\change{
We built our OpenCQA dataset based on the Chart-to-text benchmark constructed by \citet{kanthara2022chart}.} We manually went over all the charts in this benchmark to retain 7,724 that we deemed suitable for our study. In particular, we filtered out samples that are too complex and samples which we could not make an open-ended question. \Cref{chart_filtering_examples} shows examples of such samples. 

\change{In the Chart-to-text benchmark, each sample consists of a chart, the title, text extracted using an OCR method and associated summary text and the original article.  OCR text segments were extracted top to bottom left to right along with their rectangular bounding boxes given as \{x,y,len(x),len(y)\}. \change{The extracted title from the chart image was taken as the final chart title if there was no associated \texttt{alt} text with the chart image. If the \texttt{alt} text (which gives a short chart description) was available, the longer of the two was taken by comparing it with the extracted title.}
The summary for each chart was obtained by finding the most relevant texts to the chart following an annotation process which is described in detail in \cite{kanthara2022chart}.
}

\subsubsection{Data Annotation} \label{app:mturk-setup}
\begin{figure*}[!t]
     \centering
    \begin{subfigure}[b]{1\textwidth}
         \centering
         \includegraphics[width=1\textwidth,trim={0cm 0cm 0cm 5cm},clip]{ 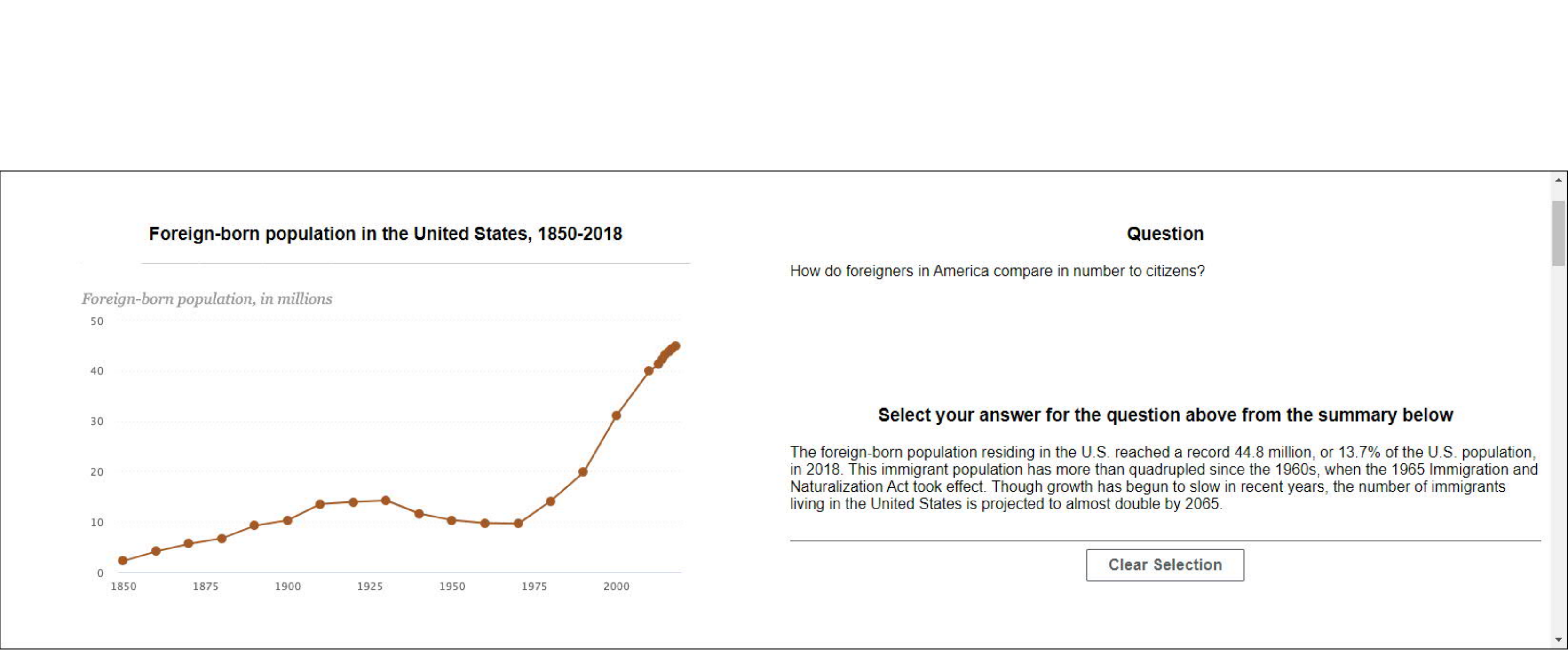}
        
         \caption{\small The interface shows a sample where the question is given and the crowdworker needs to highlight the corresponding answer from the summary.}
        \label{intf_1}
     \end{subfigure}

     \begin{subfigure}[b]{1\textwidth}
         \centering
         \includegraphics[width=1\textwidth,trim={0cm 0cm 0cm 5cm},clip]{ 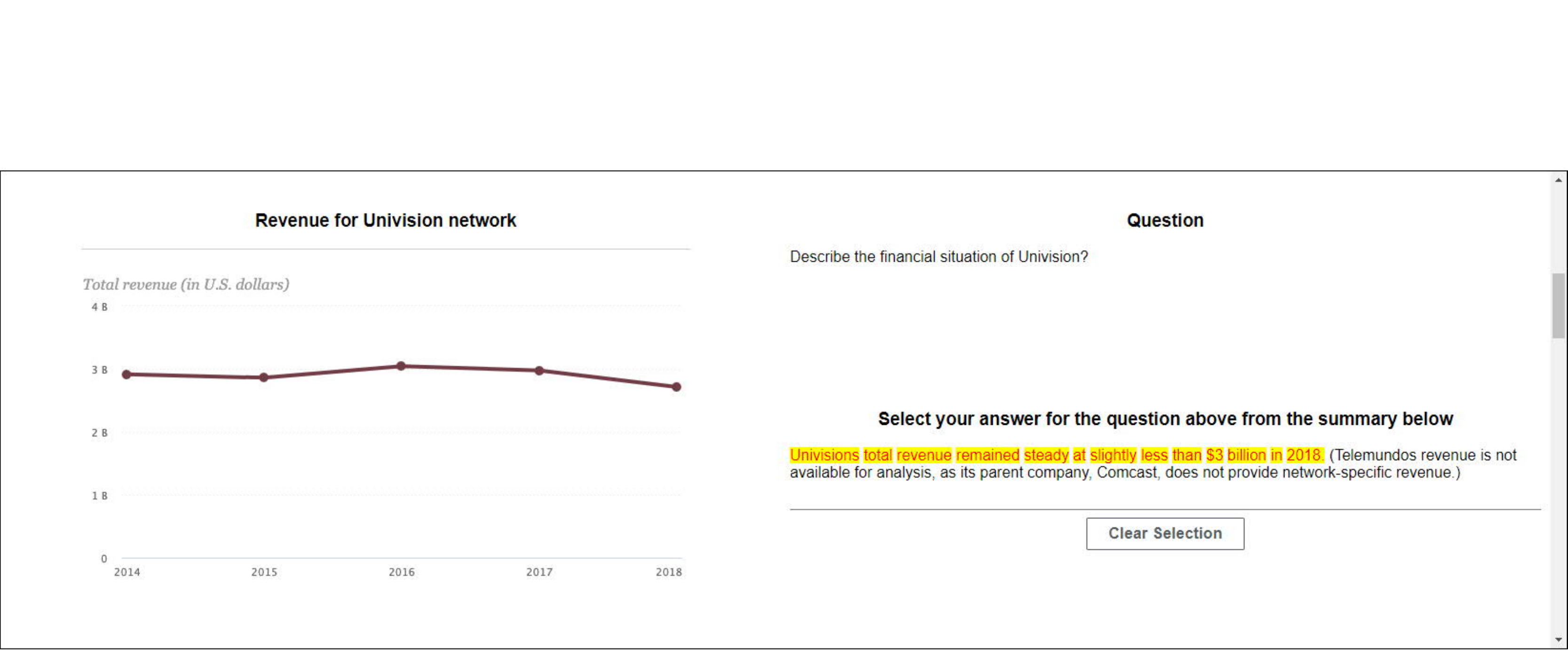}
        
         \caption{\small The interface shows the highlighted answer selected by the user for a given question.
         }
        \label{intf_2}
     \end{subfigure}
     \hfill
     \begin{subfigure}[b]{1\textwidth}
         \centering
         \includegraphics[width=1\textwidth,trim={0cm 0cm 0cm 5cm},clip]{ 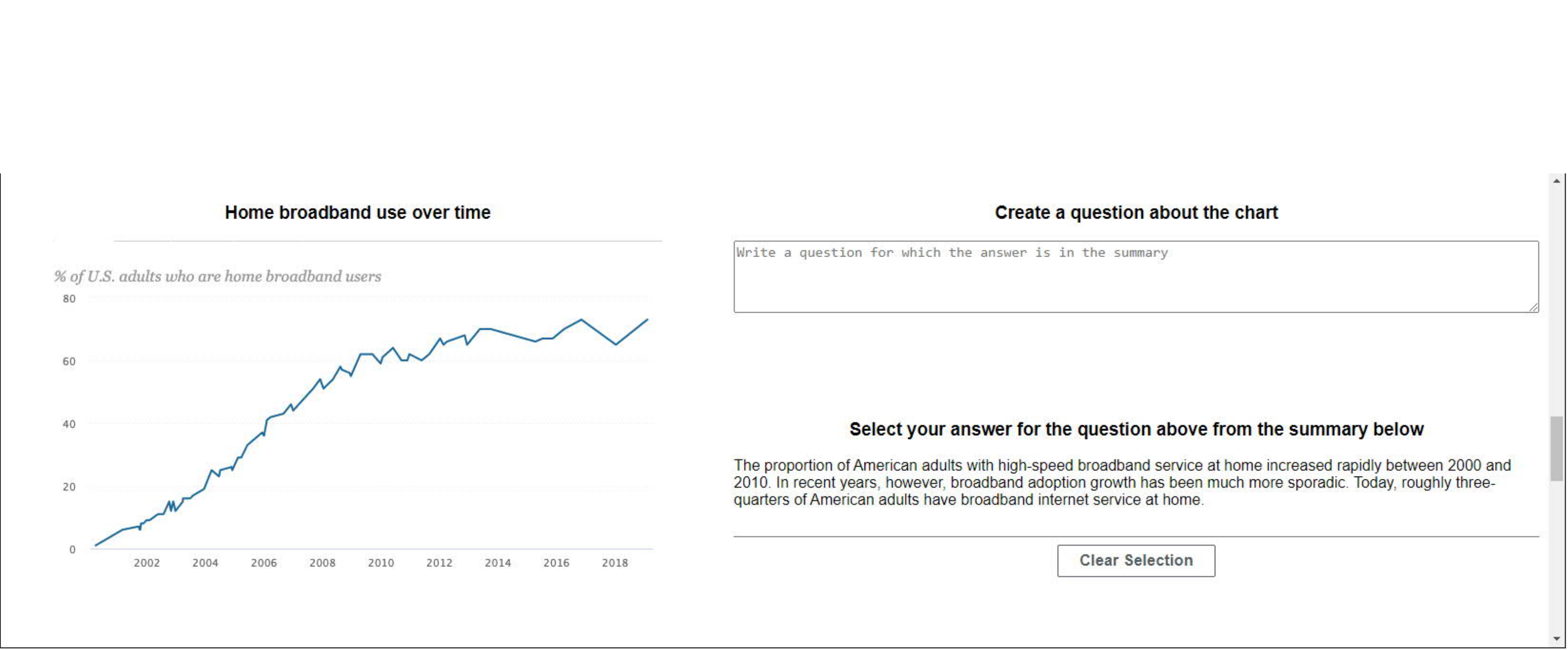}
        
         \caption{\small The interface shows an annotation task, where the annotator needs to formulate a new question and then subsequently select the answer from the summary paragraph.
         }
        \label{intf_3}
     \end{subfigure}
     \caption{\small Amazon Mechanical Turk study interface.}
    \label{mturk_interface_examples}
\end{figure*}

\change{In each HIT (Human Intelligent Task), the worker answered three questions previously asked by another coworker for three separate charts respectively and also created three new question-answer pairs for three new charts (see figure \ref{mturk_interface_examples}). The first round of 30 samples were created by the authors but were not included in the final dataset to remove any potential bias. Each subsequent round of samples were created by crowdworkers who would answer existing question that were created in the previous round while also creating new question-answer pairs for the next round.}

\begin{figure*}[!t]
     \centering
     
         \includegraphics[width=1\textwidth,trim={0cm 0cm 0cm 5cm},clip]{ 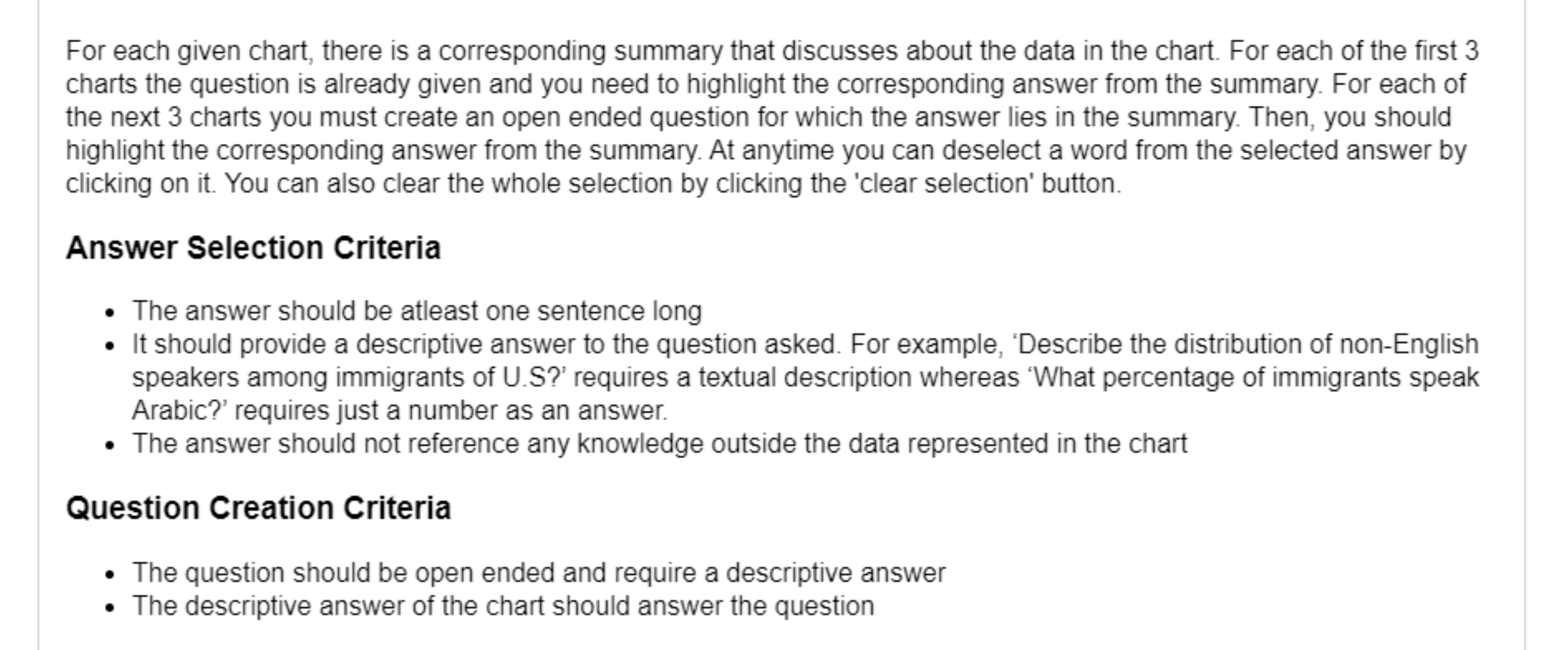}

\caption{\small Amazon Mechanical Turk Instructions for Mturk workers.}
    \label{fig:mturk_interface_instructions}
\end{figure*}

\begin{figure*}[!t]
     \centering
     
         \includegraphics[width=1\textwidth,trim={0cm 0cm 0cm 0cm},clip]{ 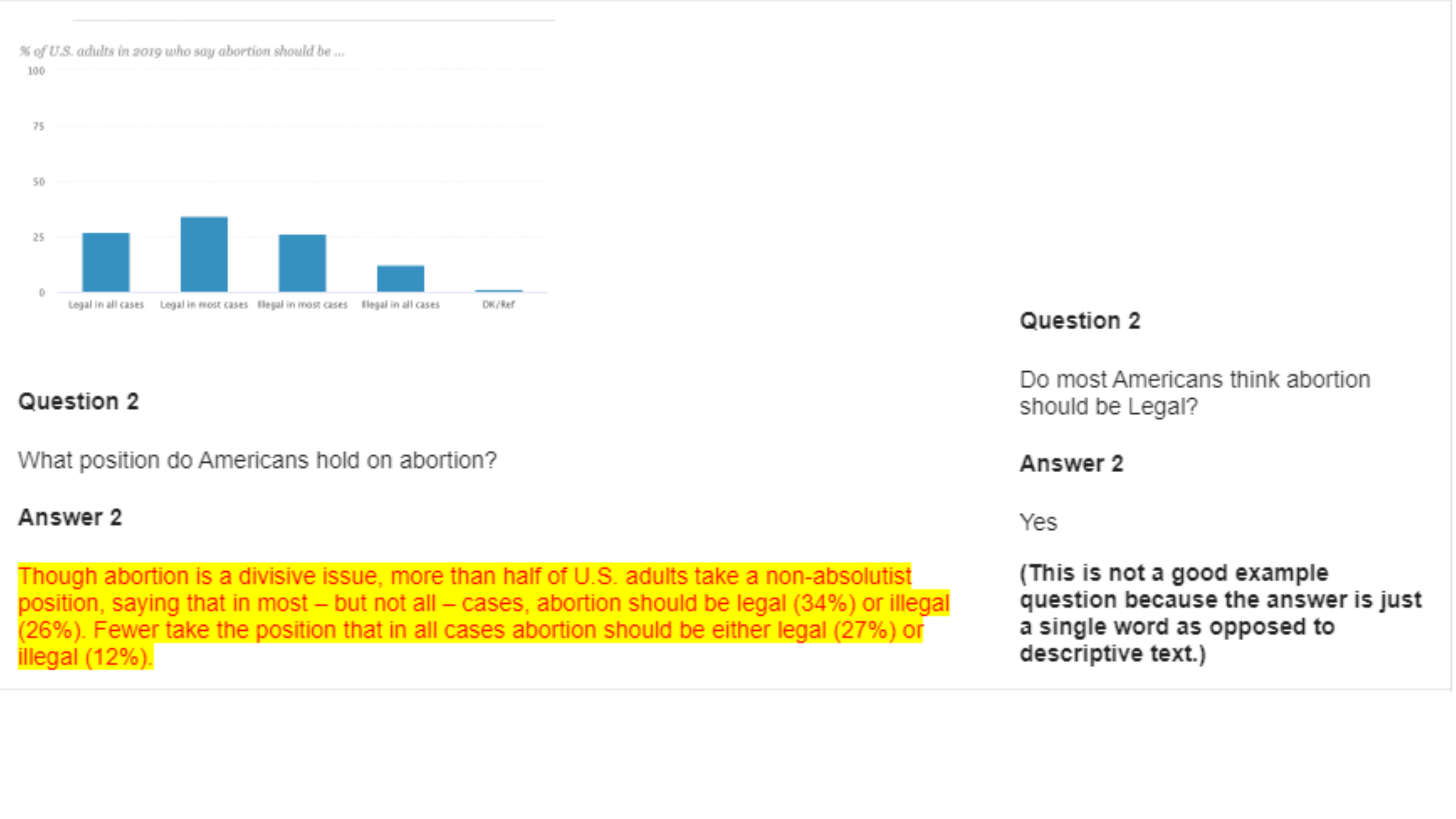}

\caption{\small Example of good and bad open ended question and answer pair on Amazon Mechanical Turk(The question on the left is a good example since it requires a descriptive answer and is open ended, and the question on the right is a bad example as it is not open ended).
}
    \label{good_n_bad_open_ended_example}
\end{figure*}

\change{Each participant was first provided with a consent form where the complete procedure of the study and data collection was explained. Once the participant provided consent, detailed instructions were shown on how to complete the task(see \Cref{fig:mturk_interface_instructions}). In particular, participants were instructed to create open-ended questions that required a descriptive answer. An answer was considered to be descriptive if it was at least one sentence long and derivable from the chart(\ie\ it did not refer to any knowledge outside the data represented in the chart). This method of selecting the answer for the question from the summary is very controlled since the summary was written by a professional writer and these summaries usually had good linguistic style and grammatical structure. Participants were additionally shown sample question-answer pairs to explain how ideal questions looked like (see \Cref{good_n_bad_open_ended_example}). }

\change{
To ensure quality, only participants with a hit approval rate greater than 95\% and over 5,000 approved hits were selected for pre-screening. In the pre-screening stage, participants completed a sample task that allowed us to assess their proficiency for this study. Those who successfully completed the task according to 
the given instructions were qualified for participating in the main study. The study  protocol was approved by the ethics review board at York University with certificate ID e2021-317.}

\begin{figure*}[h!]
\linespread{0.5}\selectfont\centering 
\renewcommand{\arraystretch}{1.0}
 \scalebox{0.6}
{\begin{tabular}{p{7cm} | p{7cm} | p{7cm}| p{5cm}}     
\toprule
        \begin{center} \vspace{-4mm} \textbf{(1) Identify} \end{center}
        \textbf{\underline{Question:}}
            What are the current thoughts on direct democracy?
            
        \raisebox{-1\height}{\hspace{-10mm}  \includegraphics[height=4cm, trim={3cm 6cm 5cm 0cm}, clip]{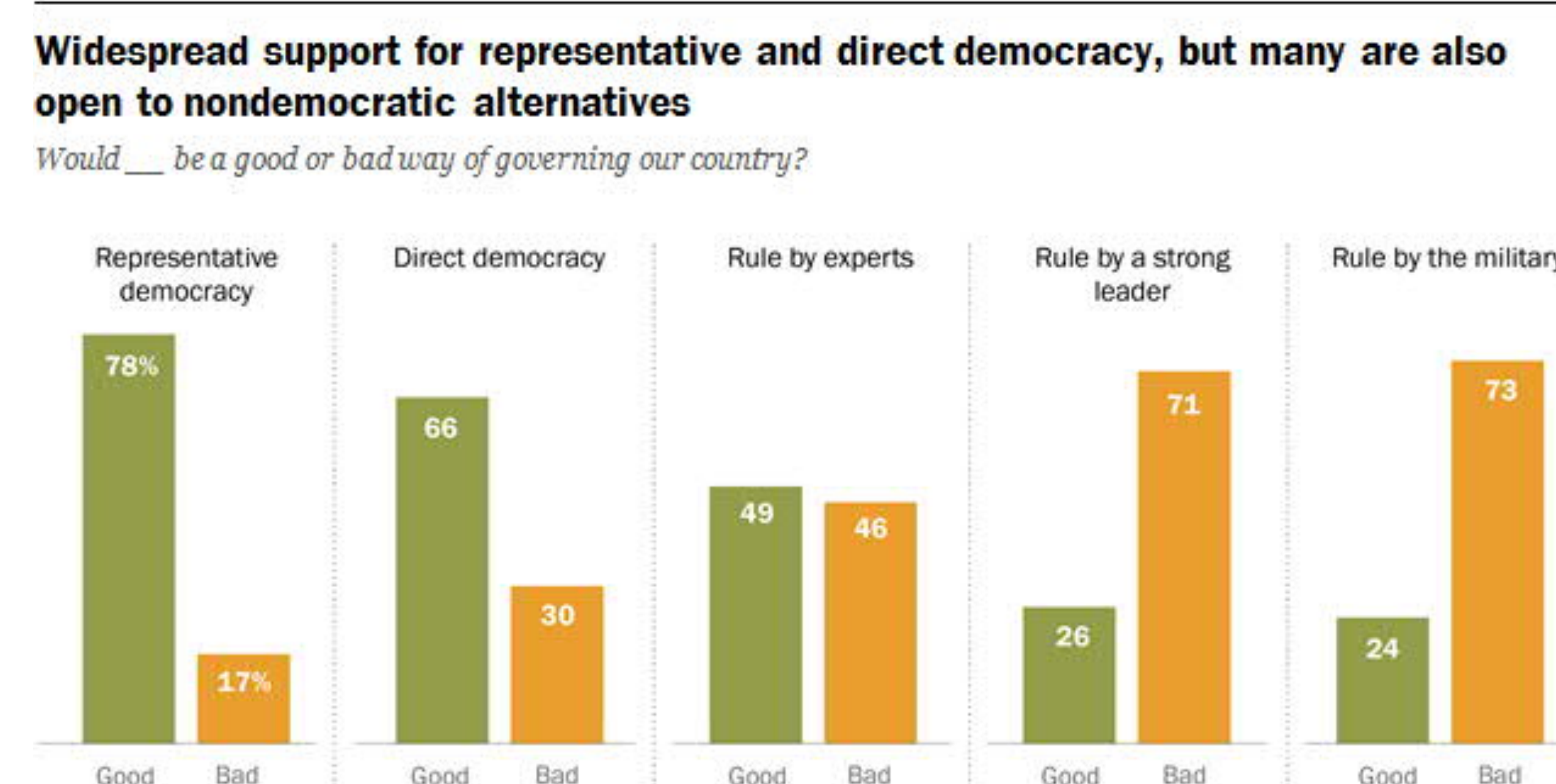}}
        &
       
        \begin{center}  \vspace{-4mm} \textbf{(2) Compare} \end{center}
        \textbf{\underline{Question:}}
            Compare Americans and Germans views about the world economic leader?

        \raisebox{-1\height}{ \hspace{-3mm}  \includegraphics[height=2cm, trim={0cm 0cm 0cm 8cm}, clip]{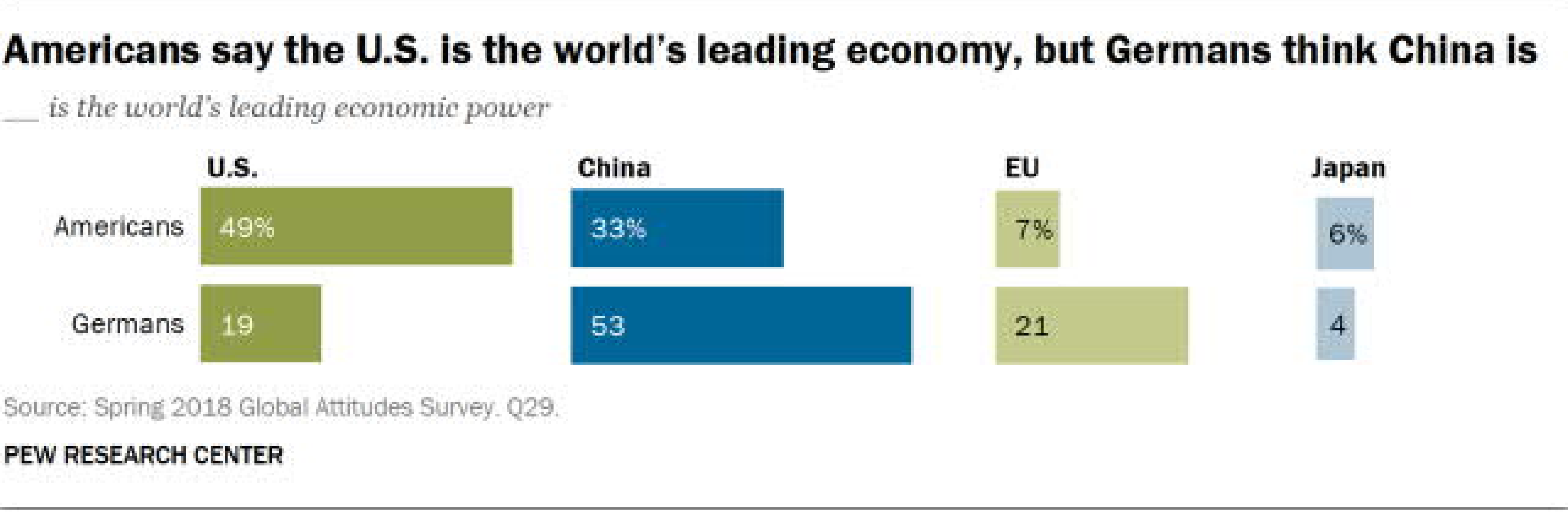}}        
        &
        \begin{center} \vspace{-4mm} \textbf{(1) Summarise} \end{center}
        \textbf{\underline{Question:}}
            Explain the distribution of people who know a transgender person?
        
        \raisebox{-1\height}{ \hspace{-3mm}  \includegraphics[height=4cm, trim={11cm 4cm 12cm 0cm}, clip]{ 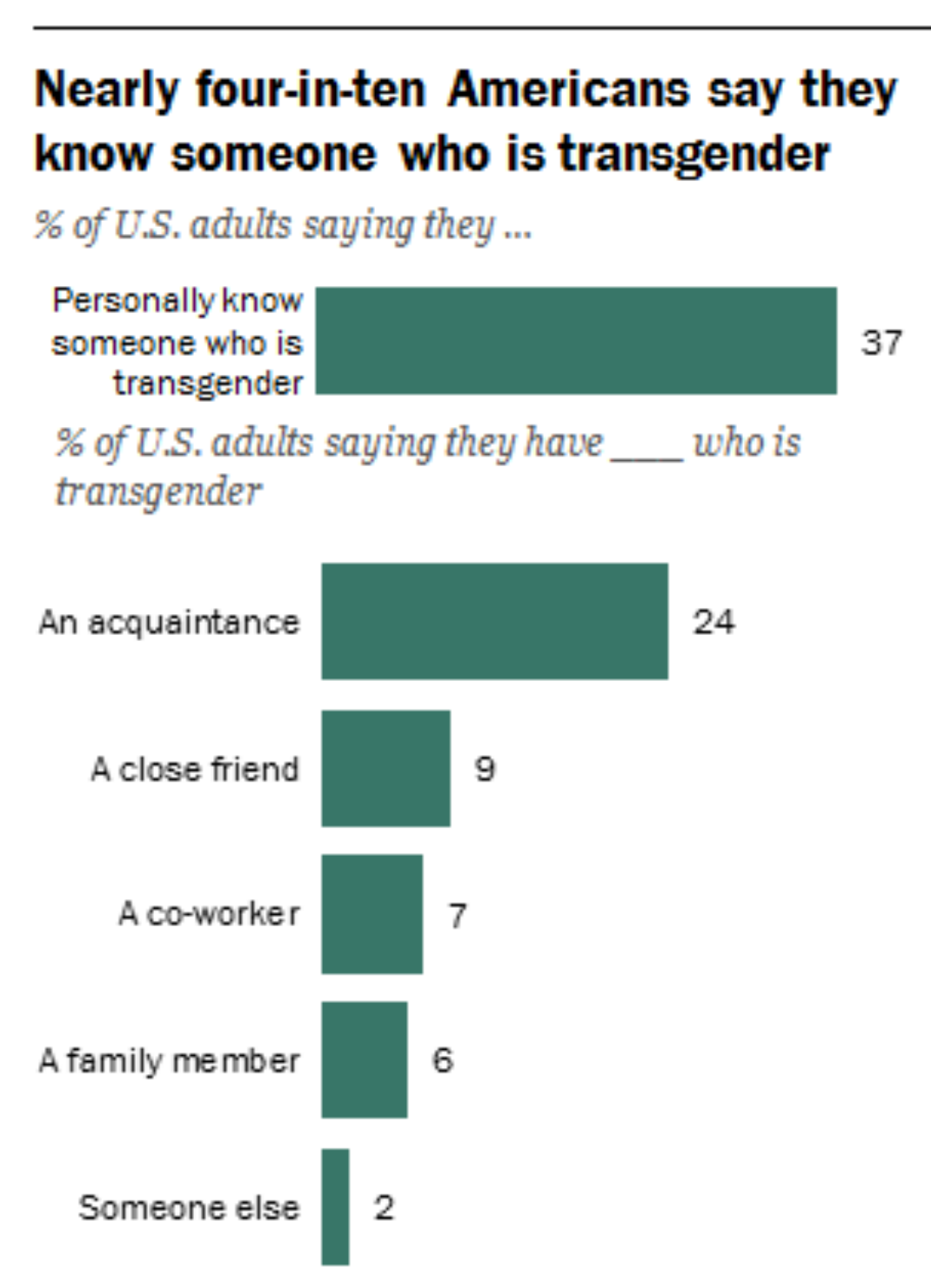}}
        
        &
       
         \begin{center} \vspace{-4mm} \textbf{(4) Discover} \end{center}
         \textbf{\underline{Question:}}
            How do Americans' see the coronavirus statistics?

        \raisebox{-1.0\height}{ \hspace{0mm} \includegraphics[height=4cm, trim={5cm 0cm 5cm 0cm}, clip]{ 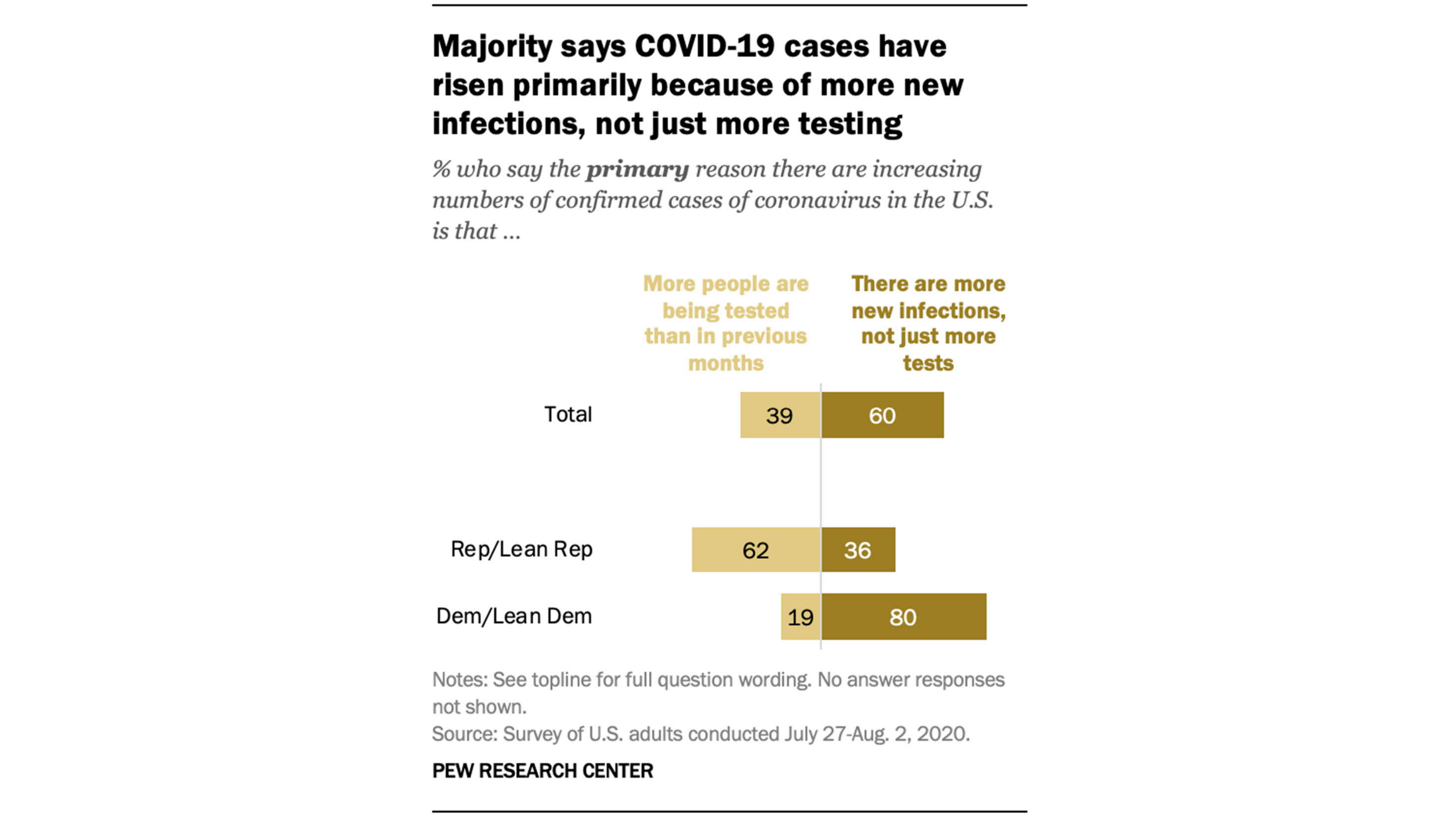}}
        \\

\midrule 

         {\small A global median of 66\% say direct democracy–in which citizens vote on major issues–would be a good way to govern.}
        &
        {\small As far as what share of Americans say they know a transgender person, 37\% say they personally do, including 13\% who say they have a close friend or a family member who is transgender (9\% say they have a close friend and 6\% have a family member who is transgender). About a quarter of Americans (24\%) say they have an acquaintance who is transgender, while 7\% say they have a transgender co-worker.}
        &
        {\small Americans and Germans diverge on who is the world ’s leading economic power. Roughly half of Americans (49\%) name the U.S. as the world ’s economic leader, while only 19\% of Germans say the same . Over half of Germans (53\%) say China is the leading economy. A further 21\% of Germans say the countries of the European Union are the world’s top economic power, while only 7\% of Americans say this.}
        &
        {\small The survey finds that a majority of Americans (60\%) say the primary reason that the number of confirmed coronavirus cases is increasing is because there are more new infections, not just more testing for the disease . About four-in-ten (39\%) say cases are rising primarily because more people are being tested than in previous months.}
        \\
        

\bottomrule  
    \end{tabular}
    }
    \vspace{-0.5em}
    \caption{\small  
    Examples question types in the Pew dataset: (1) Identify: The question refers to a specific target (`direct democracy') which represents a group of bars in the chart; (2) Compare: The question requires a comparison between two specific targets in the chart (e.g. between `Americans' and `Germans' which represent two groups of bars); (3) Summarize: The question asks to summarize the distribution statistic in the chart (i.e., `people who know a transgender person');  (4) Discover: The question does not specify any specific statistical task but requires the inference and reasoning over the whole chart to derive key insights (e.g., majority of Americans say that the increase in coronavirus cases is due to more new infections, not just because of more testing).    
    }
    \label{question_type_examples} 
    \vspace{-1mm}
\end{figure*}

\begin{figure*}[h!]
\linespread{0.5}\selectfont\centering 
\renewcommand{\arraystretch}{1.0}
 \scalebox{0.5}
{\begin{tabular}{p{10cm} | p{10cm} | p{8cm}}     
\toprule
        
        \textbf{\underline{Question:}}
            How do married adults conceive their relationship satisfaction in comparison to adults living with a partner?

        \raisebox{-1\height}{\hspace{0mm}  \includegraphics[height=7cm, trim={16cm 4cm 0cm 0cm}, clip]{ 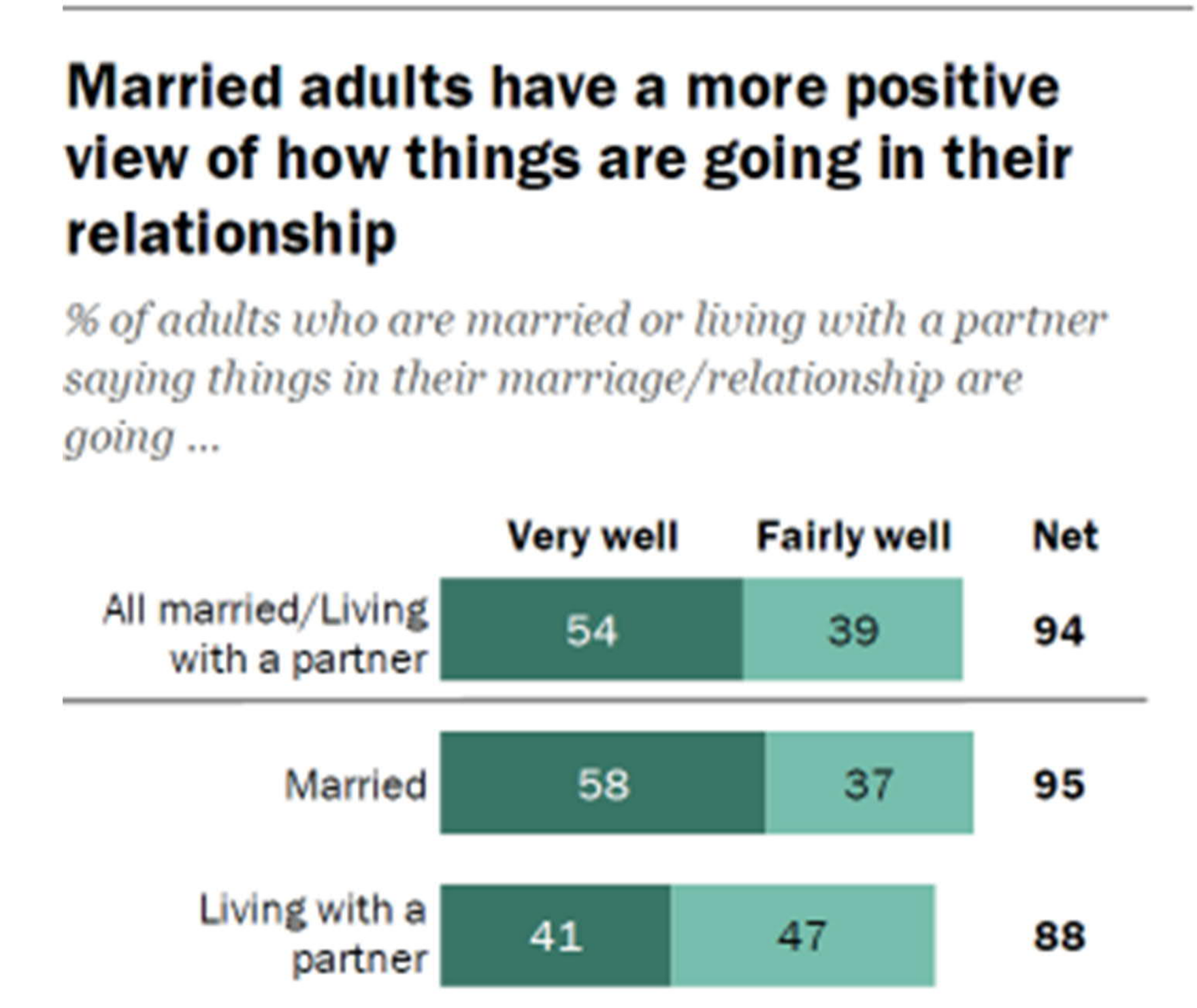}}
        
        \vspace{3mm} \textbf{\underline{VLT5:}} { Compared with adults who are married or living with a partner, many are more positive about how things are going in their relationship. Roughly nine-in-ten say that their marriage or relationship is going very or fairly well, compared with 37\% of adults who are living with a partner.}

        \vspace{2mm} \textbf{\underline{Gold Answer:}} { Married adults are more likely than those who are living with a partner to say things are going very well in their relationship (58\% vs. 41\%) They also express higher levels of satisfaction with specific aspects of their relationship, including the way household chores are divided between them and their spouse or partner how well their spouse or partner balances work and personal life, how well they and their spouse or partner communicate, and their spouse’s or partner’s approach to parenting. }

        &

        \textbf{\underline{Question:}}
            What are the partisans views on whether the marches will increase public support for science? 

        \raisebox{-1\height}{ \hspace{5mm}  \includegraphics[height=6.5cm, trim={18cm 5cm 0cm 0cm}, clip]{ 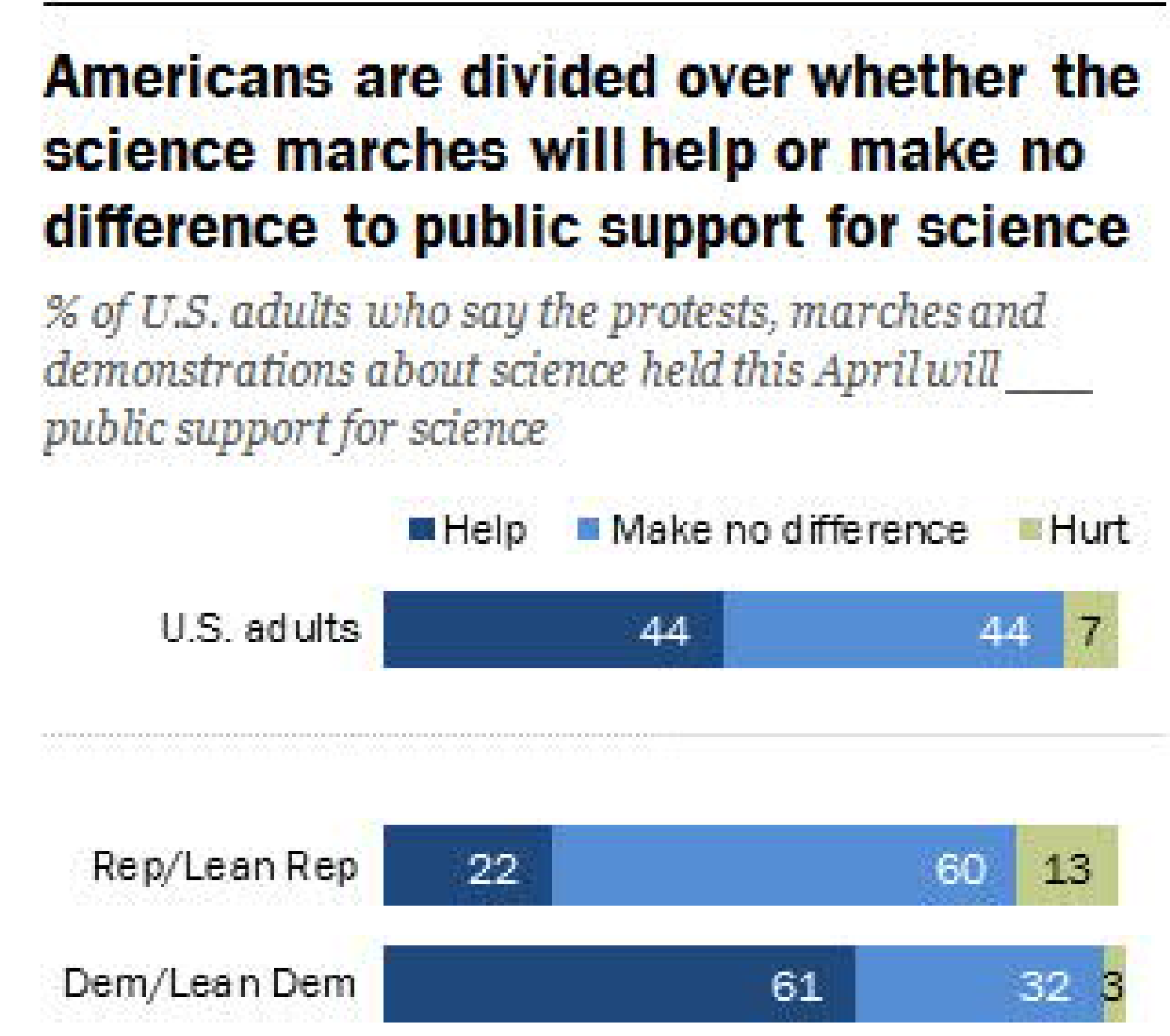}}
        
        \vspace{8mm} \textbf{\underline{VLT5:}} { Democrats and Democratic-leaning independents are divided: 60\% say the marches will increase public support for science, while just 32\% say the demonstrations will not hurt the public.}
        
       \vspace{3mm} \textbf{\underline{Gold Answer:}} { Most Democrats and Democratic leaners (61\%) believe the marches and demonstrations held in April will benefit public support for science. By contrast, just 22\% of Republicans and independents who lean to the GOP say the marches will help drive public support for science, while six-in-ten (60\%) of this group believes the marches will have no impact on public support and 13\% say the marches will hurt public support.  }
        
        &
        
        \textbf{\underline{Question:}}
            What are the public opinions about business corporations
        
        \raisebox{-1\height}{   \includegraphics[height=6.5cm, trim={15cm 3cm 0cm 0cm}, clip]{ 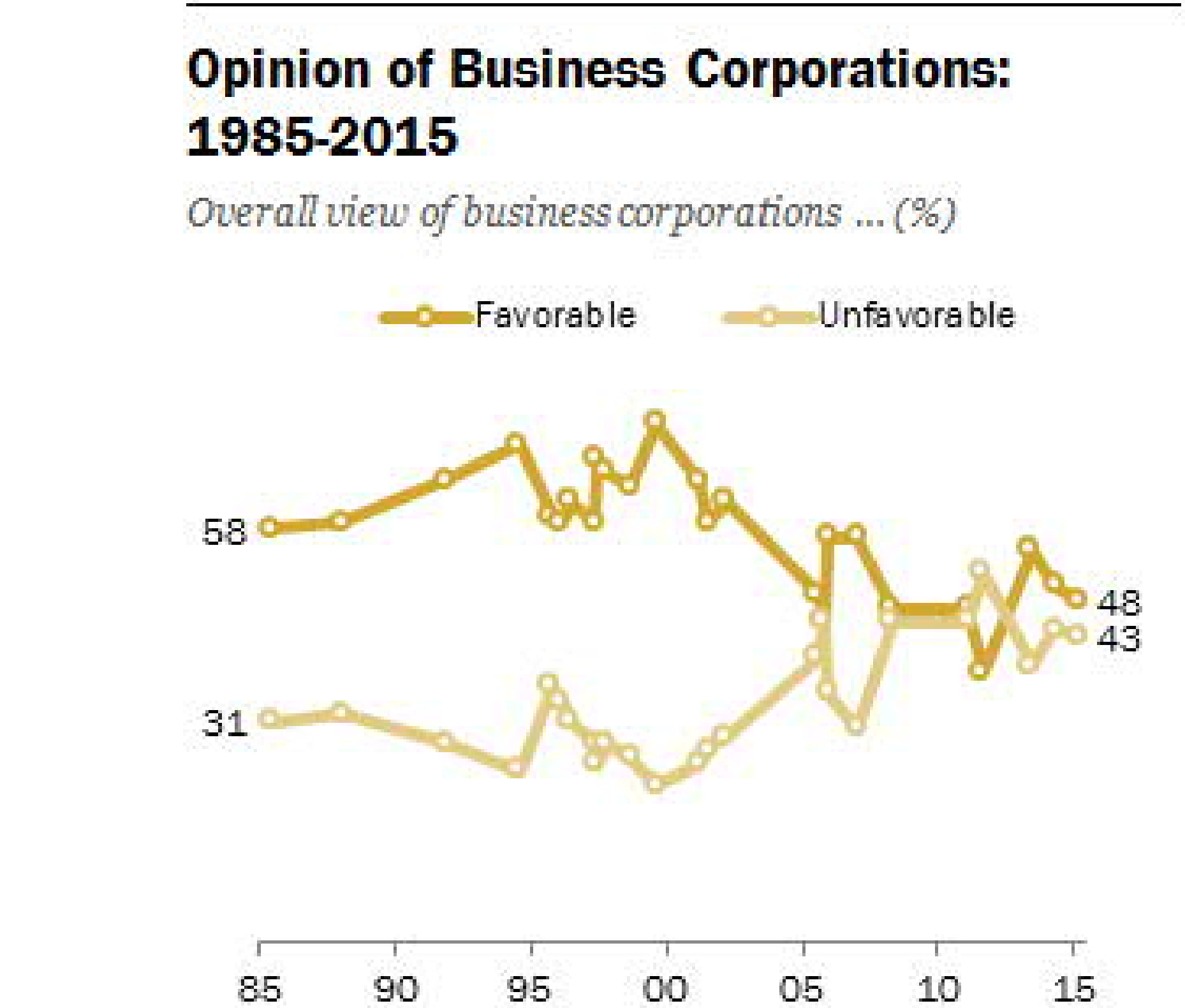}}
        
       \vspace{8mm} \textbf{\underline{VLT5:}} { Currently, 48\% express a favorable view of business corporations, while 43\% express an unfavorable view.}
        
        \vspace{5mm} \textbf{\underline{Gold Answer:}} { 48\% hold a favorable view of business corporations, compared with 43\% who hold an unfavorable view. }
        
        \\

\midrule 

        \textbf{\underline{Question:}}
            Describe the trend of Facebook users usage of the site on daily basis? 
        
        \vspace{3mm}    
        \raisebox{-1\height}{\hspace{-8mm}  \includegraphics[height=6.5cm, trim={13cm 4cm 0cm 0cm}, clip]{ 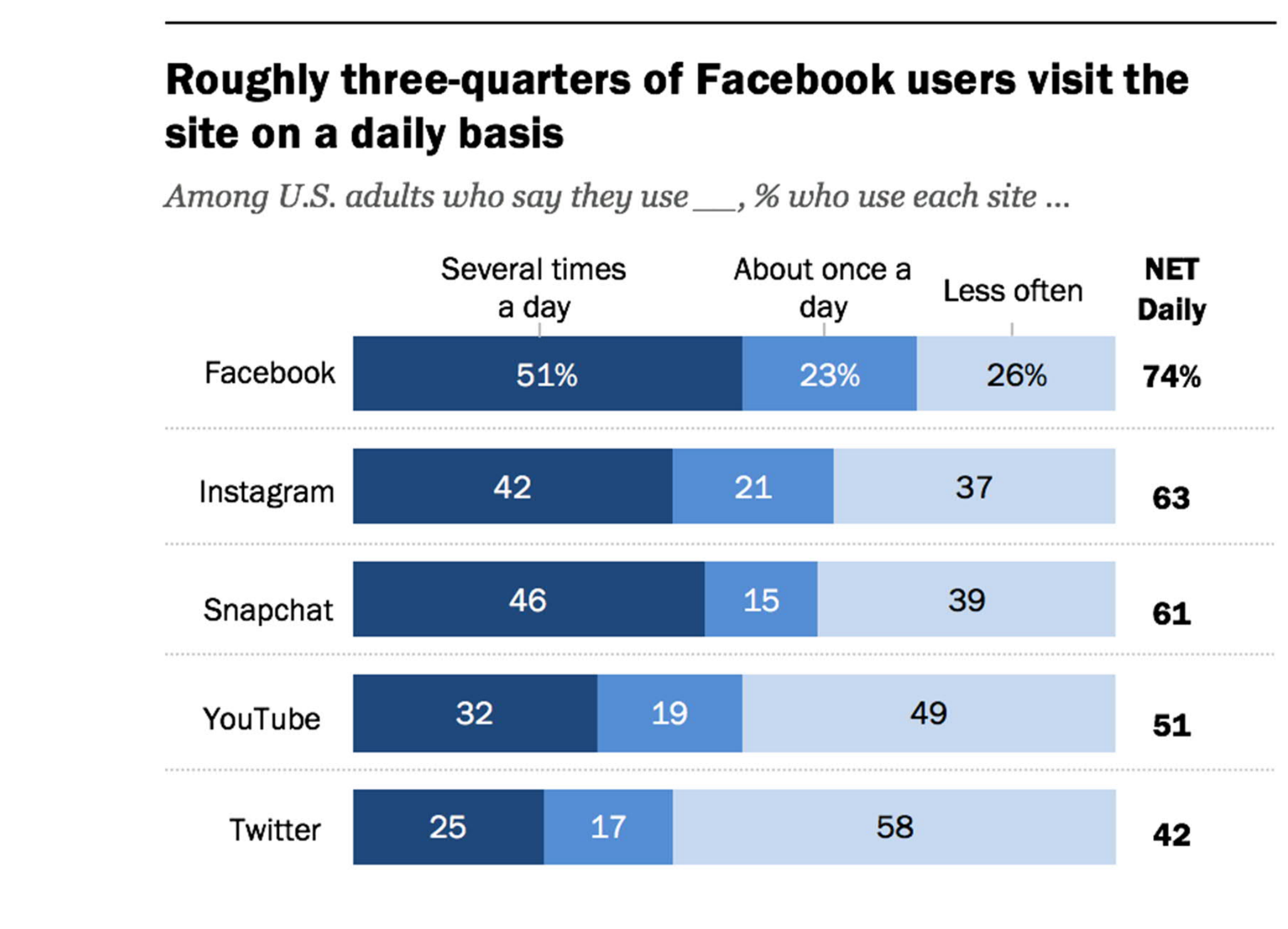}}
        
        \textbf{\underline{VLT5:}} { About three-quarters of Facebook users (71\%) say they visit the site several times a day, or several times a day. The roughly one-third (26\%) say they visit the site at least once a day, or several times a day.}
        
       \vspace{1mm} \textbf{\underline{Gold Answer:}} { Roughly three-quarters of Facebook users (74\%) visit the site daily, including about half who do so several times a day. }
        
        &

        \textbf{\underline{Question:}}
            What are views on genetic engineering of mosquitos?

        \vspace{6mm}
        \raisebox{-1\height}{ \hspace{5mm}  \includegraphics[height=6cm, trim={14cm 4cm 0cm 0cm}, clip]{ 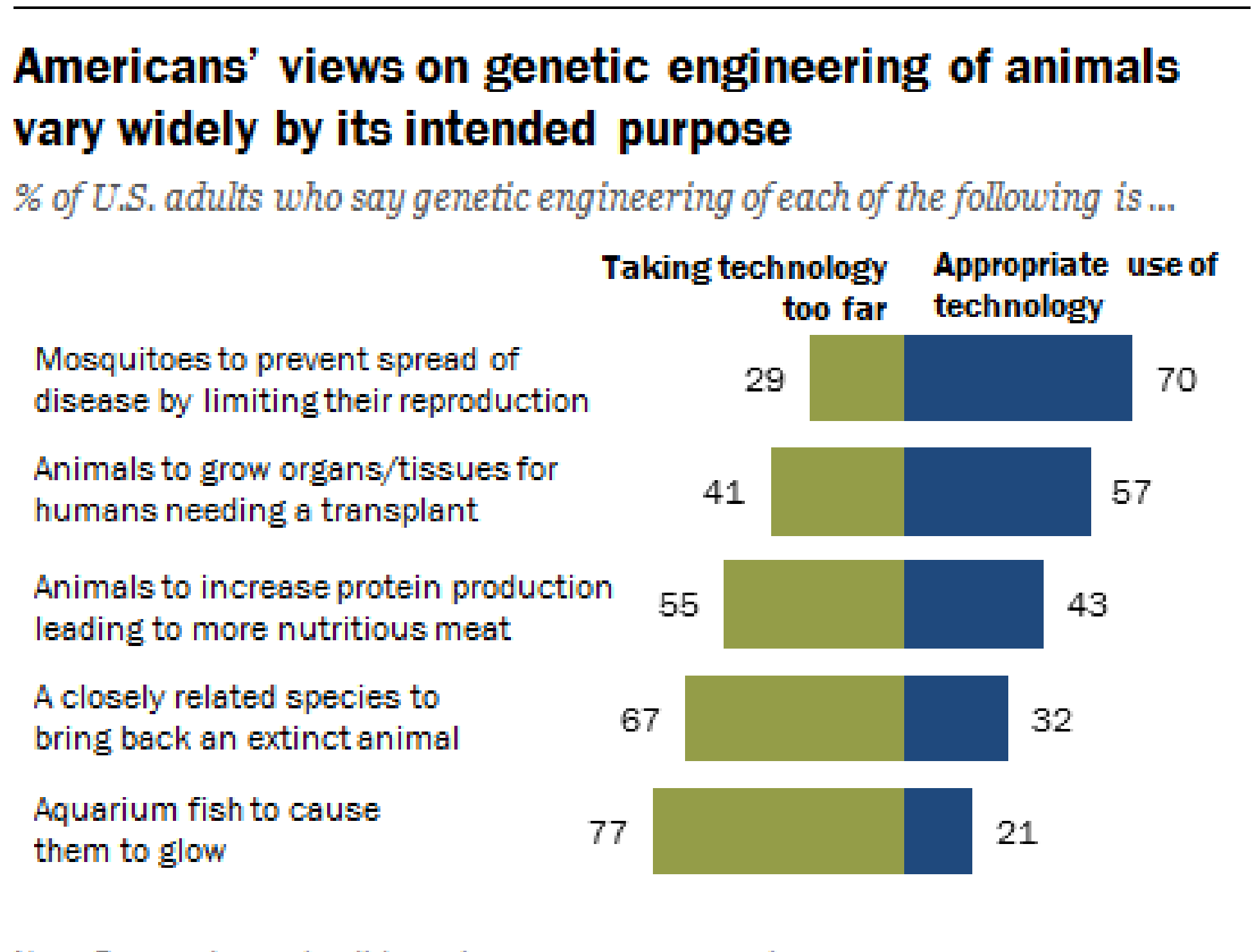}}
        
       \vspace{3mm} \textbf{\underline{VLT5:}} { Two-thirds or more of Americans (70\%) say genetic engineering of mosquitos is appropriate to prevent the spread of disease by limiting their reproduction. But a majority of Americans (55\%).}
        
       \vspace{2mm} \textbf{\underline{Gold Answer:}} { Seven -in-ten Americans (70 \%) believe that genetically engineering mosquitoes to prevent their reproduction and therefore the spread of some mosquito-borne diseases would be an appropriate use of technology, while about three-in-ten (29\%) see the use of genetic engineering for this purpose as taking technology too far.}
        
        &
        
        \textbf{\underline{Question:}}
            Compare the British approval and rejection levels about being a member of the EU.
        
        \raisebox{-1\height}{ \hspace{-3mm}  \includegraphics[height=6.5cm, trim={8cm 3cm 11cm 0cm}, clip]{ 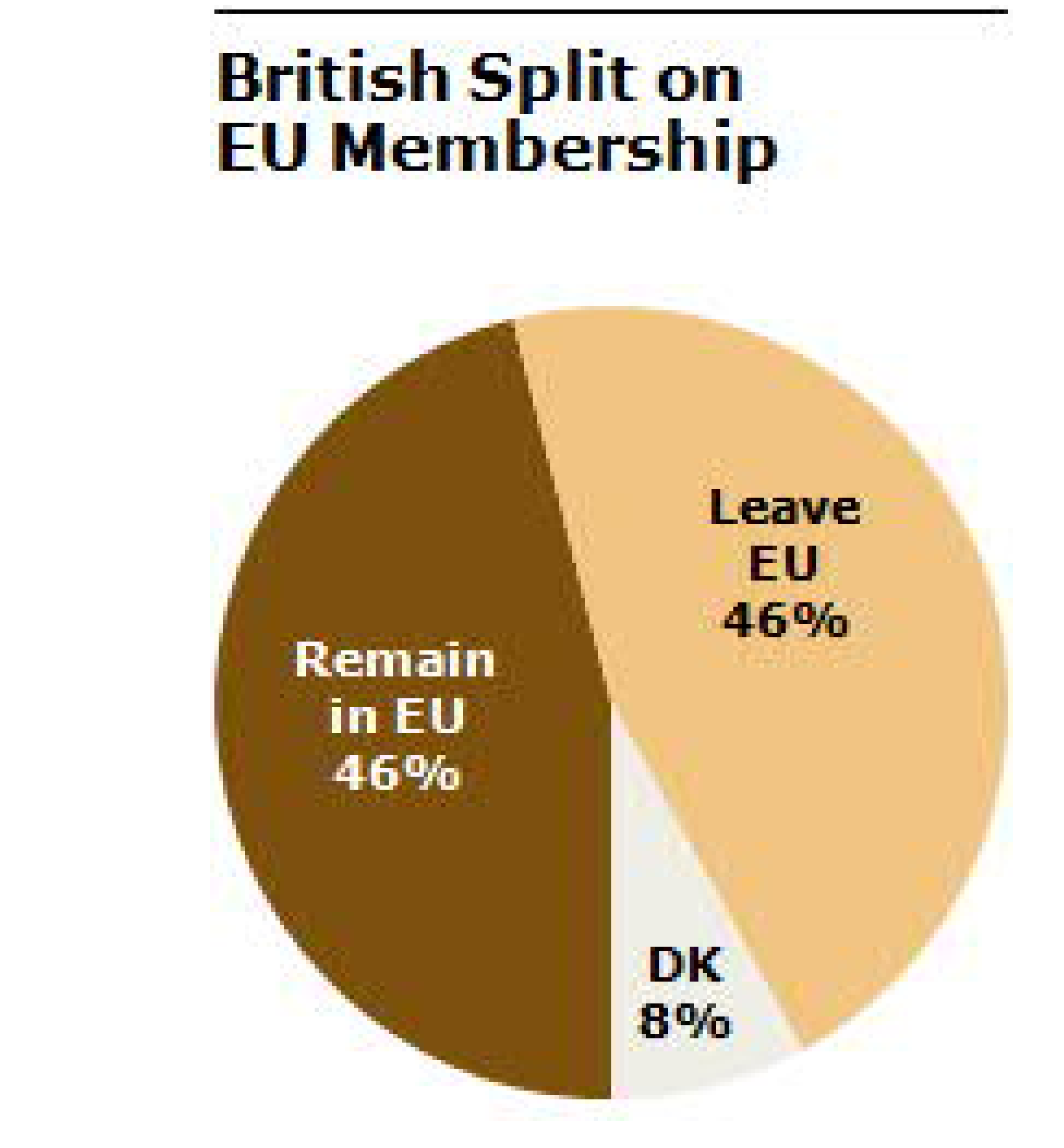}}
        
        \textbf{\underline{VLT5:}} { As the 2014-10 EU elections, Britain’s approval rating is only slightly lower than the 49\% who approved of becoming a member while only 8\% said they would leave.}
    
        \vspace{2mm} \textbf{\underline{Gold Answer:}} { The British, remain evenly divided on leaving the EU: 46\% say stay and 46\% say go. }
        \\

\bottomrule  
    \end{tabular}
    }
    \vspace{-0.5em}
    \caption{\small  
    Sample outputs from the best performing model VLT5 (without summary). 
    }
    \label{model_output_samples} 
\end{figure*}

\begin{figure*}[!t]
     \centering
    \begin{subfigure}[b]{1\textwidth}
         \centering
         \includegraphics[width=1\textwidth,trim={0cm 0cm 0cm 0cm},clip]{ 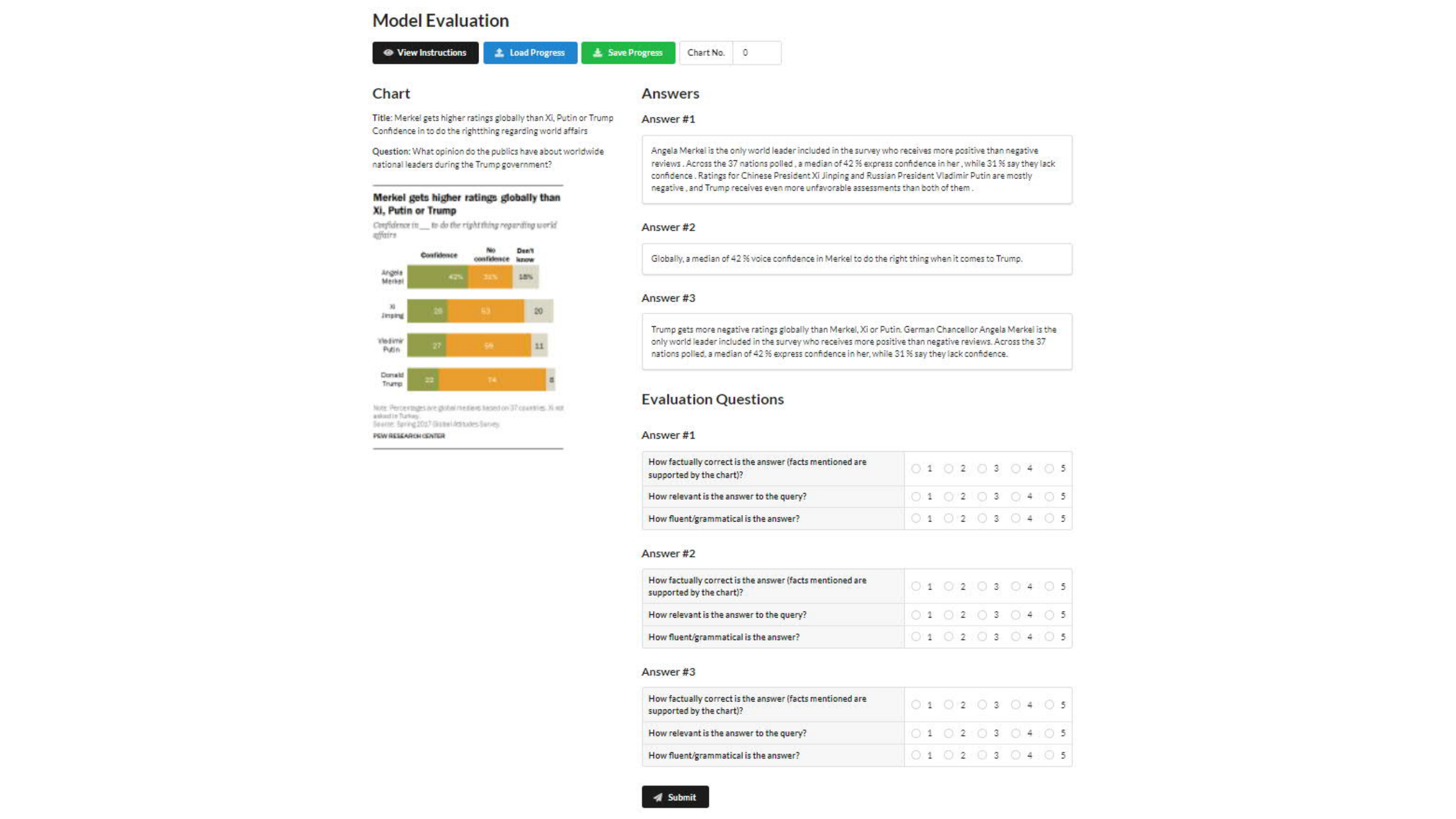}
        
         \caption{\small 
         The interface for human evaluation that shows the gold answer as well as the outputs from  VLT5 (VLT5 without summary) and VLT5-S(VLT5 with summary) in random order. The factuality, relevance and fluency measures are rated on a 5-point likert scale.
         }
        \label{intf_1}
     \end{subfigure}

     \begin{subfigure}[b]{1\textwidth}
         \centering
         \includegraphics[width=1\textwidth,trim={0cm 0cm 0cm 0cm},clip]{ 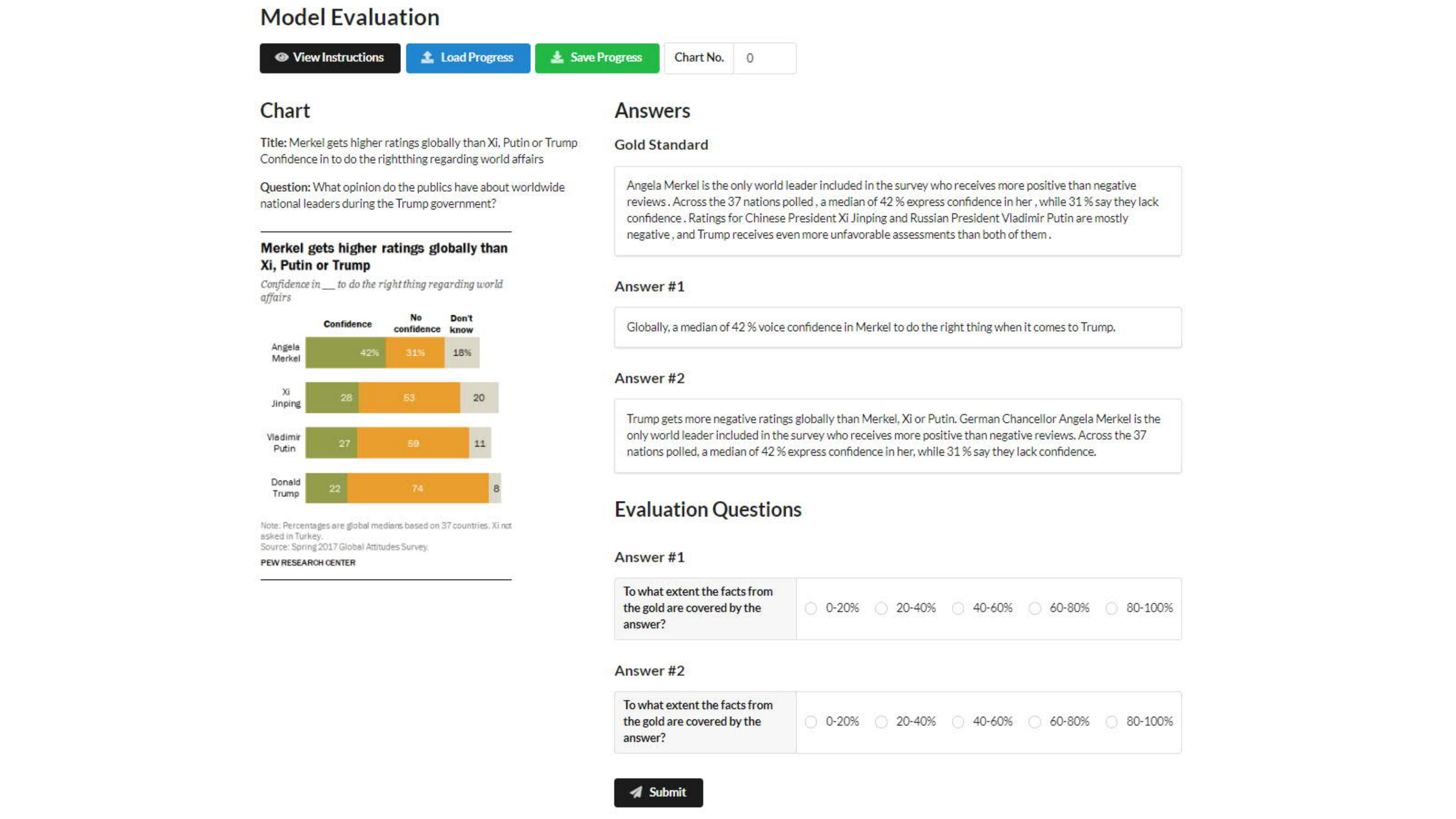}
        
         \caption{\small 
The interface for human evaluation on the content seleciton measure. The human compares between the gold answer and the outputs from VLT5 (VLT5 without summary) and VLT5-S(VLT5 with summary) presented in random order.
         }
        \label{intf_2}
     \end{subfigure}
     \hfill
     \caption{\small Human evaluation interfaces.}
    \label{human_evaluation_interface}
\end{figure*}


\subsection{Baseline Models}
\label{app:baselines}


The architectures of different example models along with sample inputs and outputs are shown in Figure \ref{sample_models}. \change{All models concatenate the question, title and OCR extracted data and additionally the summary/article based on the problem setting. 
}We provide more details on the hyperparameter settings for different models below.

\paragraph{GPT2} \change{We fine-tune GPT2 with} a batch size of 6 for 5 epochs with a max input sequence length of 512 tokens and output sequence length of 128 tokens.  

\paragraph{BERTQA} \change{We fine-tune} BERT-large ( 340M parameters, 24 layers) with a batch size of 4 for 2 epochs, with a max input sequence length of 512 tokens and output sequence length of 128 tokens.

\paragraph{CODR}
\change{We fine-tune BART-large} for 10 epochs and a batch size of 6 with max input sequence length of 512 tokens, output sequence length of 128 tokens,gradient accumulation set to 1 and the other parameters at default settings.

\paragraph{BART} We fine-tune BART-Base\footnote{\url{huggingface.co/transformers}\label{footnote:hugging}} (140M, 6-layers) with max input sequence length of 512 tokens,
for 50K iterations with a batch size of 4 and evaluate after every 1,000 iterations on the validation set. The initial learning rate is set to 0.00005. For inference, we use the model with the lowest validation loss and decode with a beam size of 4.

\paragraph{T5} Similar to BART, we fine-tune T5-Base\footref{footnote:hugging} (220M, 12-layer Transformer as the encoder and decoder) with max input sequence length of 512 tokens, for 100k iterations with a batch size of 4 and an initial learning rate of 0.00001, evaluate after every 1,000 iterations on validation set, and use {the model with best validation loss for testing. Inference is done with beam search with a beam size 4.} 

\vspace{-7mm}
\change{\paragraph{VLT5} {We fine-tune the pretrained VLT5 (T5-base as its backbone) for 100 epoches with a learning rate of 0.0001 and max input sequence length of 512 tokens. Inference is done with beam search with a beam size of 4.} 
Note that the  original VL-T5 represents an input image with 36 object regions. However, since the number of bounding-box objects in charts varies from one to another, we either pad the extracted visual features (with zeros) or truncate them to maintain the length at 36.}


\paragraph{Electra} We fine-tune Electra-Base 
for 25,000 iterations with a batch size of 4 and initial learning of 0.0001. The max sequence size of input is set to \change{2048}. For inference, the batch size is set to 4 with search beam size of 20.

\begin{table*}[t!]
    \centering
    \small
    \scalebox{0.9}
    {\begin{tabular}{lccccccc}
        \toprule
        \textbf{Models} & \textbf{Type} & \change{\textbf{ROUGE\_-F (1/2/L)}} $\uparrow{}$ & \textbf{CS} $\uparrow{}$ & \textbf{BLEURT} $\uparrow{}$ & \textbf{CIDEr} $\uparrow{}$ & \change{\textbf{BERTScore}} $\uparrow{}$ & \textbf{BLEU} $\uparrow{}$\\
        
        \midrule
        \multicolumn{7}{c}{\change{\textbf{Setup 1:With Article Provided}}}  \\
   \rowcolor{gray}     Bboxes-BERTQA & Extractive & 23.95/9.17/16.69 & 32.71\% & -0.736 & 1.080 & 85.10 & 9.35 \\
   \rowcolor{gray}     Bboxes-ELECTRA & Extractive & 52.12/43.07/47.49 &  49.74 \% &  -0.051  &   4.379  & 89.97 & 30.77 \\ 
        Bboxes-BART & Generative & 45.07/31.93/39.55 & 48.48\% & -0.108 & 3.411 & 90.20 & 16.78   \\
        Bboxes-T5 & Generative & 66.21/60.15/63.40 & 67.49\% & 0.076 & 5.414 & \textbf{93.55} & \textbf{41.58} \\
        
        \change{BBoxes-VLT5} & Generative & 55.91/46.53/51.76 & 62.05\% & 0.048 & 4.932 & 91.83 & 37.94 \\
        Bboxes-GPT2 & Generative & 16.78/6.89/14.08 & 15.45\% &  -0.757 & 0.85 & 81.12 & 1.36 \\
        Bboxes-CODR & Generative & 13.67/0.97/10.88 & 1.73\% & -1.026 & 0.035 & 82.40 & 0.00 \\
        
        
    \rowcolor{gray}    BERTQA & Extractive  & 22.81/8.53/16.63 & 28.17\% & -0.692 & 0.983 & 85.00 & 9.58 \\
    \rowcolor{gray}    ELECTRA & Extractive  & 57.67/50.09/53.89 &  54.73 \% &  0.066 &   5.100  & 91.01 & 38.79 \\
        BART & Generative  & 50.68/38.95/45.55 & 54.29\% & -0.017 & 4.121 & 91.08 &  23.91 \\
        T5 & Generative  & 66.57/60.40/\textbf{63.46} & \textbf{68.24\%} & \textbf{0.103} & \textbf{5.437} & 93.53 & 41.28 \\
         \change{VLT5} & Generative  & 58.90/49.97/54.82 & 66.06\% & 0.076 & 5.227 & 92.34 & 40.06 \\
        GPT2 & Generative  & 16.71/66.57/13.75 & 14.71\% & -0.745 & 0.814 & 82.05 & 0.99 \\
        CODR & Generative  & 14.58/1.31/11.37 & 3.4\% & -1.155 & 0.051 & 81.90 & 0.43 \\

        \midrule
        \multicolumn{7}{c}{\textbf{Setup 2:With Summary Provided}}  \\
   \rowcolor{gray}     Bboxes-BERTQA & Extractive & 71.76/68.56/70.13 & 85.25\% & 0.309 & 7.216 & 94.73 & \textbf{67.24} \\
   \rowcolor{gray}     Bboxes-ELECTRA & Extractive & 75.87/73.78/75.09 & 91.93\% & 0.367 & 7.766 & 95.98 &  65.17 \\ 
        Bboxes-BART & Generative & 63.11/57.22/60.44 & 65.93\% & 0.205 & 6.090 & 93.23 & 36.12 \\
        Bboxes-T5 & Generative & 76.54/74.24/75.35 & 84.21\% & 0.377 & 7.780 & 95.36 & 60.26 \\
        
        \change{BBoxes-VLT5} & Generative & 73.91/71.07/72.67 & 80.92\% & 0.341 & 7.470 & 94.98 & 58.25 \\
        Bboxes-GPT2 & Generative & 18.21/15.41/17.47 & 25.73\% & -0.686 & 1.955 & 43.28 & 16.06  \\
        Bboxes-CODR & Generative & 12.16/0.73/9.31 & 2.040\% & -1.283 & 0.046 & 80.17 & 0.19 \\
        
        
   \rowcolor{gray}     BERTQA & Extractive  & 70.97/67.77/69.34 & 85.28\% & 0.297 & 7.177 & 94.65 & 66.33 \\
    \rowcolor{gray}   ELECTRA & Extractive  & 76.24/74.33/\textbf{75.48} & \textbf{92.57\%} & \textbf{0.378} &  \textbf{7.823} & \textbf{96.06} &  65.40 \\ 
        BART & Generative  & 66.41/61.86/64.41 & 68.71\% & 0.257 & 6.613 & 93.85 &  38.42 \\
       
        T5 & Generative  & 75.77/73.51/74.51 & 81.73\% & 0.369 & 7.728 & 95.22 & 57.93 \\
      
        \change{VLT5} & Generative  & 75.07/72.35/73.86 & 85.36\% & 0.376 & 7.597 & 95.20 & 59.80 \\
        
        GPT2 & Generative  & 14.29/11.13/13.42 & 22.00\% & -0.782 & 1.725 & 49.18 & 12.68 \\
       
        CODR & Generative  & 13.81/0.76/10.25 & 2.480\% & -1.039 & 0.038 & 81.87 & 0.31 \\

        \midrule
        \multicolumn{7}{c}{\textbf{Setup 3:Without Summary Provided}} \\
        Bboxes-BART & Generative & 39.79/21.24/31.94 & 49.12\% & -0.171 & 2.254 & 89.63 &  7.40 \\
        BBoxes-T5 & Generative & 40.88/21.86/32.96 & 51.12\% & -0.181 & 2.339 & 89.56 &  9.09 \\
        
        \change{BBoxes-VLT5} & Generative & 42.62/22.18/32.93 & 53.26\% & \textbf{-0.132} & 2.382 & 89.52 & 14.01 \\
        Bboxes-GPT2 & Generative & 28.57/10.91/22.24 & 31.30\% & -0.459 & 1.246 & 84.50 & 3.92 \\
        Bboxes-CODR & Generative & 14.18/1.19/10.22 & 4.43\% & -1.053 & 0.053 & 81.77 & 0.22 \\
        
         BART & Generative & 40.29/21.40/32.48 & 49.07\% & -0.166 & 2.260 & \textbf{89.69} & 7.41 \\
        T5 & Generative  & 41.12/22.09/32.97 & 52.30\% & -0.173 & 2.357 & 89.59 & 9.28\\
        \change{VLT5} & Generative  & 42.87/22.60/\textbf{33.29} & \textbf{54.47\%} & -0.134 & \textbf{2.447} & 89.53 & \textbf{14.73} \\
        GPT2 & Generative  & 28.55/11.26/22.46 & 32.00\% & -0.493 & 1.314 & 85.05 & 4.89 \\
       CODR & Generative  & 14.67/1.05/10.90 & 4.14\% & -1.170 & 0.053 & 81.86 & 0.32 \\
        
        \bottomrule 
    \end{tabular}}
    \vspace{-0.5em}
    \caption{\small 
    Ablation study results for understanding the effect of including the OCR extracted bounding boxes as input. All models use OCR-extracted data as input.  
    "Bboxes-" models use bounding box information of chart elements. 
    \vspace{-4mm}
    } 
\label{tab:ablation-table} 

\end{table*}

\subsection{Additional Results from Evaluation}
\label{app:additional}

\subsubsection{Performance by Chart Types}
We analyze the performance by chart types  in~\Cref{tab:results_charttype}. We observe that our best performing models perform better at summarizing simple (pie charts) and frequent chart types (bar charts), whereas the performance generally decreases for  \change{complex and less frequent charts (\eg scatter plots)}.

\subsubsection{Ablation Test}
\label{app:ablation}

In table \ref{tab:ablation-table} we show the results of an ablation test, where each model is  given the question and OCR-extracted chart data as input. \change{Here, models with "Bboxes" prefix are given the bounding boxes $C_{bbox}$ of text segments in the chart, and the other models are repeated here from \Cref{tab:evaluation-table} for comparison. We observes that the inclusion of the bounding box information does not necessarily improves the performance. One possible reason could be that bounding box extraction of the OCR model is not always perfect as the model sometimes merges or splits the bounding boxes incorrectly. Note that for ROUGE-F evaluation we used the python package rouge-score, which is a Google re-implementation of the original Rouge-1.5.5 perl script.}




\begin{table}[t!]
    \centering
    \small
    \scalebox{0.8}
    {\begin{tabular}{lccc}
    \toprule
     \textbf{Models} & \textbf{CEF-Precision} $\uparrow{}$ & \textbf{CEF-Recall} $\uparrow{}$ & \textbf{CEF-Coverage} $\uparrow{}$  \\
    \midrule
    \textbf{Test}     & 49.32 & 25.89\% & 47.34  \\
    \midrule
        \multicolumn{4}{c}{\textbf{Setup 1:With Article Provided}} \\
        \textbf{Bboxes-VLT5}     & 38.05\% & 21.16\% &  42.18\% \\
        \textbf{VLT5}     & 42.32\% & 22.57\% &  43.11\% \\
        \textbf{Bboxes-T5}     & \textbf{42.82\%} & 22.63\% &  43.60\% \\
        \textbf{T5}     & 42.50\% & \textbf{24.31\%} &  \textbf{45.30\%} \\
    \midrule
        \multicolumn{4}{c}{\textbf{Setup 2:With Summary Provided}} \\
        \textbf{Bboxes-VLT5}     & 42.94\% & 25.18\% &  46.71\% \\
        \textbf{VLT5}     & \textbf{43.48\%} & 25.15\% &  46.96\% \\
        \textbf{Bboxes-T5}     & 42.53\% & 25.78\% &  47.74\% \\
        \textbf{T5}     & 42.04\% & \textbf{26.96\%} &  \textbf{49.06\%} \\
    \midrule
        \multicolumn{4}{c}{\textbf{Setup 3:Without Summary Provided}} \\
        \textbf{Bboxes-VLT5}     & 48.70\% & 22.46\% &  43.54\% \\
        \textbf{VLT5}     & 50.18\% & 23.09\% &  \textbf{44.08\%} \\
        \textbf{Bboxes-T5}     & 60.71\% & 20.36\% &  38.50\% \\
        \textbf{T5}     & \textbf{61.89\%} & \textbf{21.24\%} &  40.32\% \\
    \bottomrule
    \end{tabular}}
    \caption{ \small Chart Extraction Factor (CEF) scores for VLT5 and T5 across all problem setups. "Bboxes-" models use bounding box information of chart elements.     
     }
    \label{tab:cef}
\end{table}


\begin{table*}[t!]
    \centering
    \small
    {\begin{tabular}{lcccccc}
    \toprule
     \textbf{Models} & \textbf{ROUGE\_-F (1/2/L)} $\uparrow{}$ & \textbf{CS} $\uparrow{}$ & \textbf{BLEURT} $\uparrow{}$ & \textbf{CIDEr} $\uparrow{}$ & \textbf{BERTScore} $\uparrow{}$ & \textbf{BLEU} $\uparrow{}$ \\
    \midrule
    \textbf{T5-Only Question}     & 31.57/13.19/25.05 & 31.66\% &  -0.396 &  1.401 &  87.94 & 5.34 \\
    \bf{T5-Only Summary} &  67.56/63.78/65.47 & \textbf{82.22\%} & 0.199 &\textbf{6.856} & 93.94 & 49.18 \\
    \textbf{VLT5-Only Question}     & 29.56/11.56/22.43 & 32.98\% &  -0.363 &  1.322 &  87.74 & 6.34 \\
    \textbf{VLT5-Only Summary}     & 67.31/63.54/\textbf{65.79} & 77.47\% &  \textbf{0.238} & 6.803 &  \textbf{94.07} & \textbf{52.71} \\
   \textbf{Random Summary Baseline}  & 54.48/46.22/49.07 & 62.28\% & -0.015 & 4.997 & 90.89 & 45.69 \\
   \textbf{Random Article Baseline}  & 23.95/8.81/17.12 & 19.66\% & -0.641 & 0.960 & 84.36 & 5.50 \\
    \bottomrule
    \end{tabular}}
    \caption{\small Results for T5 and VLT5 models with only the question or the summary as input on OpenCQA test set and a baseline that randomly selects 52\% and 7\% tokens for the with summary and with article setups respectively. }
\label{tab:add_exps}

\end{table*}


\begin{table*}[t!]
    \centering
    \small
    {\begin{tabular}{lccccc}
    \toprule
   \multicolumn{5}{l}{\textbf{\change{VLT5 (Without Summary)}} }  \\
    \midrule
    & \bf{Bar} & \bf{Line} & \bf{Pie} & \bf{Area} & \bf{Scatter} \\
    \midrule
    \change{\textbf{ROUGE}} $\uparrow{}$     & 43.75/23.41/33.96 &   39.86/19.82/30.64 &    48.18/26.12/39.12 &  100.0/100.0/100.0&   31.89/9.21/19.33 \\
    \bf{BLEURT} $\uparrow{}$ &  -0.134 & -0.150 & -0.020 & 0.800 & -0.329 \\
    \bf{CIDEr} $\uparrow{}$ & 2.576 & 2.058 & 2.912 & 0.0 & 1.181 \\
    \change{\bf{BERTScore}} $\uparrow{}$ & 89.60 & 89.21 & 90.86 & 98.79 & 86.35\\
    
    \midrule
   \multicolumn{5}{l}{\textbf{\change{VLT5 (With Summary)}} } \\
    \midrule
    & \bf{Bar} & \bf{Line} & \bf{Pie} & \bf{Area}  & \bf{Scatter}\\
    \midrule
    \change{\textbf{ROUGE}} $\uparrow{}$  &  75.61/72.9/74.38 & 73.19/70.46/71.93 &  80.1/77.57/79.48 &100.0/100.0/100.0 &  67.46/66.83/67.46 \\
    \bf{BLEURT} $\uparrow{}$ & 0.381&0.353&0.457&0.800&0.302 \\
    \bf{CIDEr} $\uparrow{}$ & 7.657&7.414&7.920&0.0&7.378 \\
    \change{\bf{BERTScore}} $\uparrow{}$ & 95.28&94.87&96.24&98.79&93.47\\
    \bottomrule
    \end{tabular}
    }
    \vspace{-2mm}
    \caption{Results of VLT5 (without summary) and VLT5 (with summary) on our test set \wrt chart types.
    }
    \label{tab:results_charttype}
\vspace{-2mm}
\end{table*}

\begin{table*}[t!]
\centering
\small 
\resizebox{\linewidth}{!}
{
\begin{tabular}{p{5cm}p{7.5cm}p{4.8cm}}
\toprule
\textbf{Question Validation \& Editing} & \textbf{Disagreement Resolution} & \textbf{Decontextualization} \\
\midrule

Describe the \textcolor{applegreen}{concerns on} mobile phone\textcolor{applegreen}{s}  \textcolor{cadmiumred}{\st{addictions}}  \textcolor{applegreen}{among the surveyed countries}?

\raisebox{-1\height}{\includegraphics[scale = 0.3,trim={7.8cm 3.4cm 0cm 2.8cm},clip]{ 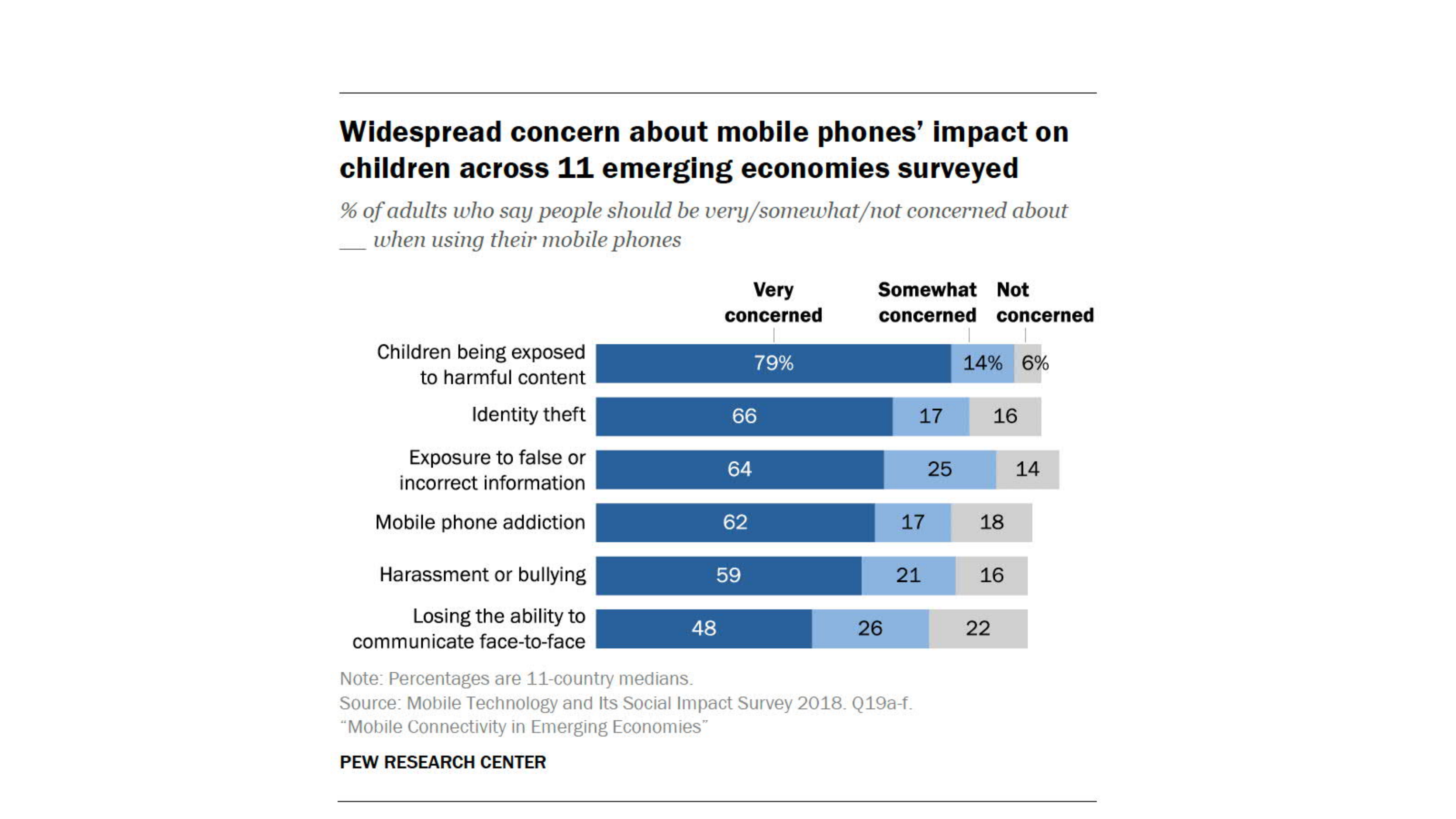}}

&

\textbf{Ans 1:} \textcolor{darkorchid}{In addition to their concerns about} \textcolor{brown}{the impact of mobile phones on children, majorities across the 11 countries surveyed also say people should also be very worried about issues such as identity theft (an 11-country median of 66\% say people should be very concerned about this),} \textcolor{darkorchid}{exposure to false information (64\%)} \textcolor{brown}{addiction (62\%) and harassment or bullying (59\%) when using their mobile phones} \textcolor{darkorchid}{face-to-face due to mobile phone use (48\%)} \newline

Ans 2: \textcolor{brown}{the impact of mobile phones on children, majorities across the 11 countries surveyed also say people should also be very worried about issues such as identity theft (an 11-country median of 66 \% say people should be very concerned about this)}\textcolor{azure}{, mobile phone} \textcolor{brown}{addiction (62\%) and harassment or bullying (59\%) when using their mobile phones}

&   

In addition to their concerns about the impact of mobile phones on children, majorities across the 11 countries surveyed \textcolor{cadmiumred}{\st{also}} say people should \textcolor{cadmiumred}{\st{also}} be very worried about issues such as identity theft (an 11-country median of 66\% say people should be very concerned about this), exposure to false information (64\%)\textcolor{applegreen}{, mobile phone} addiction (62\%) and harassment or bullying (59\%) when using their mobile phones\textcolor{applegreen}{. Fewer are very concerned about the risk that people might lose the ability to communicate} face-to-face due to mobile phone use (48\%). \\[-0.5em]
\bottomrule
\end{tabular}
}
\vspace{-3mm}
\caption{
    \small Illustration of data  annotation process (steps 2 - 4). \textcolor{applegreen}{Green} text is newly added \change{from the summary}, \textcolor{cadmiumred}{\st{strikeout red}} text is removed. \textcolor{purple}{Purple} text is from the first worker, \textcolor{azure}{blue} text is from the second worker, and \textcolor{brown}{brown} represents texts in common between the participants. 
}
\label{analysis-process}
\vspace{-3mm}
\end{table*}

\begin{table*}
\centering
\small 
\resizebox{\linewidth}{!}
{
\begin{tabular}{p{7cm} | p{7cm} | p{4cm}}
\toprule

 \multicolumn{2}{c}{\textbf{Model Input}}  & \textbf{Generated Answer} \\
\midrule

\textbf{Question:}
What are the partisans views on whether the marches will increase public support for science? \newline


\raisebox{-1\height}{\includegraphics[scale = 0.55,trim={10.8cm 0cm 9cm 5.5cm},clip]{ 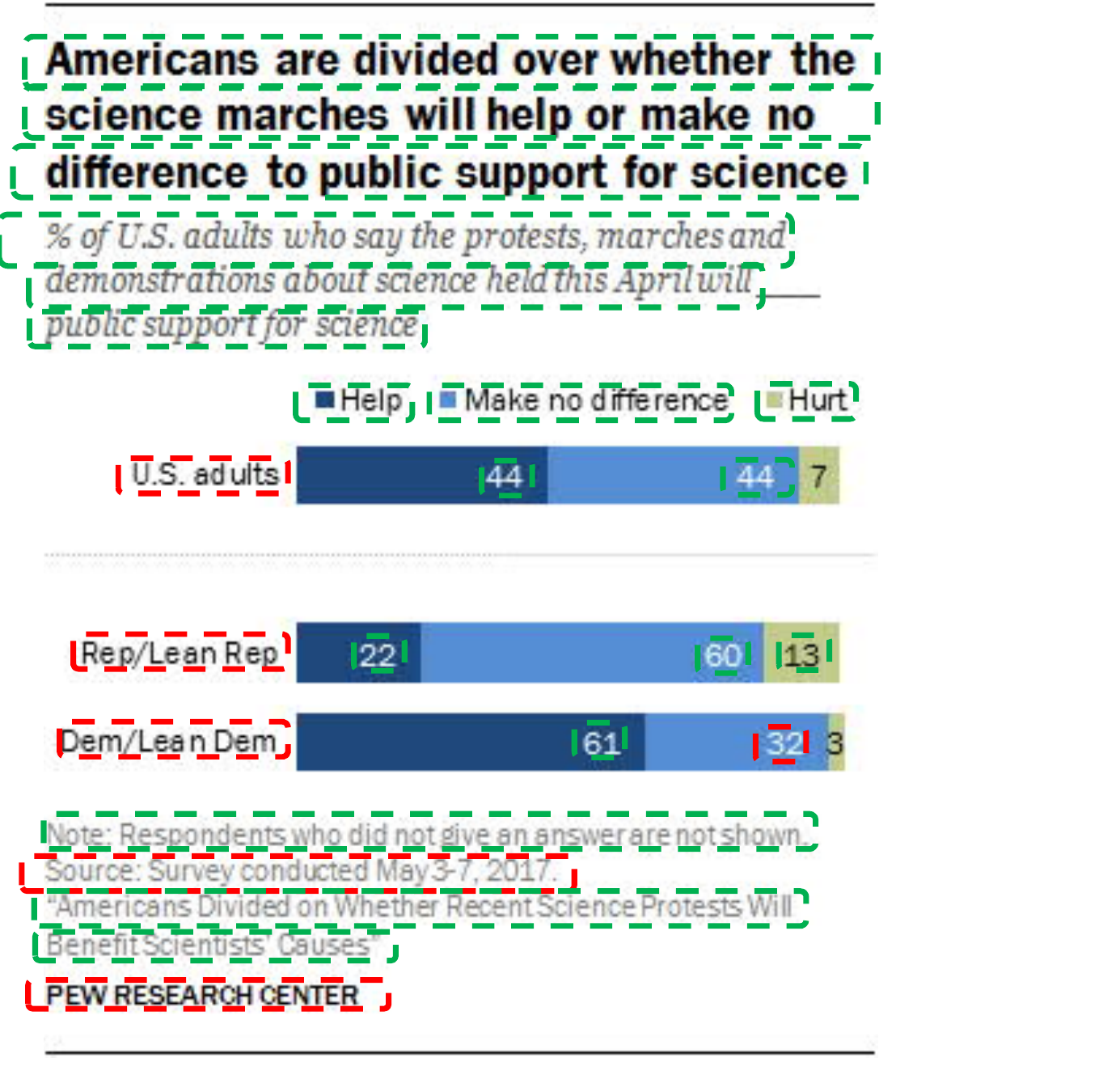}}

&


\textbf{OCR extracted data:} \textcolor{cadmiumred}{\textcolor{applegreen}{Americans are divided} orever \textcolor{applegreen}{whether the} | \textcolor{applegreen}{science marches will help} or make \textcolor{applegreen}{no} | \textcolor{applegreen}{difference} topublic \textcolor{applegreen}{support for science} | \textcolor{applegreen}{\% of} USS adults who say \textcolor{applegreen}{the} protests, \textcolor{applegreen}{marches} Aprilwill \textcolor{applegreen}{and} | \textcolor{applegreen}{demonstrations public} about \textcolor{applegreen}{science} held this Aprilwill | \textcolor{applegreen}{public} supportfor \textcolor{applegreen}{science} | \textcolor{applegreen}{Help Make no} d ifference \textcolor{applegreen}{Hurt} | U.S ad ults | \textcolor{applegreen}{44} | \textcolor{applegreen}{44} | Rep/Lean Rep | \textcolor{applegreen}{22} | 60 | \textcolor{applegreen}{13} | Dem/Lean Dem | \textcolor{applegreen}{61} | 32 | Note: Respondents who did notgive 2017 an answerare.not.shown. | "Americans Source Survey \textcolor{applegreen}{Divided} conducted May3-7 notgive Recent 2017 | "Americans Survey \textcolor{applegreen}{Divided} \textcolor{applegreen}{on Whether} Recent \textcolor{applegreen}{Science} Protests \textcolor{applegreen}{Will} | \textcolor{applegreen}{Benefit} Scientists' Ca uses | PEW RESEARCH CENTER} \newline


\textbf{Summary:} Americans are \textcolor{azure}{closely} divided on the \textcolor{azure}{issue} of whether the marches will \textcolor{azure}{increase} public support for science: 44\% \textcolor{azure}{think they} will help and 44\% \textcolor{azure}{think they} will make no difference. \textcolor{azure}{Another 7\% believe they} will hurt public support . \textcolor{azure}{Most Democrats and Democratic leaners} (61\% ) \textcolor{azure}{believe} the marches and demonstrations held in \textcolor{azure}{April} will benefit public support for science. By \textcolor{azure}{contrast}, just 22\% of \textcolor{azure}{Republicans and independents who lean to the GOP} say the marches will help \textcolor{azure}{drive} public support for science, while \textcolor{azure}{six-in-ten} (60\%) of this \textcolor{azure}{group believes} the marches will have no \textcolor{azure}{impact} on public support and 13\% say the marches will hurt public support. \newline

&   


\textcolor{applegreen}{Americans are \textcolor{azure}{closely} divided on the \textcolor{azure}{issue} of whether the marches will \textcolor{azure}{increase} public support for science: 44\% \textcolor{azure}{think they} will help and 44\% think they will make no difference. \textcolor{azure}{Another 7\% believe they} will hurt public support. \textcolor{azure}{Most Democrats and Democratic leaners} (61\%) \textcolor{azure}{believe} the marches and demonstrations held in \textcolor{azure}{April} will benefit public support for science. By \textcolor{azure}{contrast, just} 22\% of \textcolor{azure}{Republicans} and \textcolor{azure}{independents who lean to} the \textcolor{azure}{GOP} say the marches will help \textcolor{azure}{drive} public support for science , while \textcolor{azure}{six-in-ten} (60\%) of this \textcolor{azure}{group believes} the marches will have no \textcolor{azure}{impact} on public support and 13\% say the marches will.} \\

\bottomrule
\end{tabular}
}
\vspace{-1mm}
\caption{
    \small Illustration of chart extraction scoring.\textcolor{applegreen}{Green} text are answer tokens generated from OCR text, \textcolor{azure}{Blue} text are answer tokens not found in OCR text but generated from summary, \textcolor{cadmiumred}{Red} text highlights the remaining OCR-extracted text. All text sequences in the chart with bounding boxes are extracted by the OCR model and \textit{Green} bounding boxes are used in the generated answer. CEF(Recall): \textbf{43.48}; CEF(Coverage): \textbf{86.36}; CEF(Precision): \textbf{54.54}.
}
\label{fig:cef_example}

\end{table*}

\subsubsection{Evaluation with Naive Baselines}
\change{
In order to better understand the dataset and how challenging the task is, we analyze the results of \change{fine-tuning and }testing T5 and VLT5, our best performing models with only \textit{the question} $Q$ or \textit{the summary} $S$ as input (see \Cref{tab:add_exps})
In addition, we  choose a naive baseline which randomly selects the answer from the given input text. This baseline spans 52\% tokens of the input when only the summary $S$ is given  whereas it spans  7\% tokens of the input when the article $D$ is given (based on the statistics of tokens overlapping between the \change{summary or article} and the gold answer in our dataset(see \Cref{dataset_stats_n_ctypes})). 
As expected, we can see that both models struggle to answer the question when only the question is given as input. However, they perform better with only summary as input as the model has access to the summary containing the answer. In contrast, the performance drops when we use the baseline that randomly selects the answer, especially when the entire article is given as input as it becomes more difficult to randomly guess the answer.}
\subsubsection{The Effect of the OCR-text} 
\change{We further analyze how the model extracts the answer from different input sources (e.g., input summary vs. OCR-generated text). For this purpose, we consider three new metrics for Chart Extraction Factor (CEF) scores as follow:


\noindent \textbf{CEF-Precision:} Shows 
  how much fraction of generated tokens in the answer (excluding stopwords) $A$ is coming from OCR-extracted text segments $C_{label}$.
  

\noindent \textbf{CEF-Coverage:} Represents how much fraction of OCR-extracted text segments $C_{label}$ contribute to the generated answer $A$. An OCR-extracted text segment contributes to the answer if any generated tokens in the answer $A$ appear in the segment.  
  
\noindent \textbf{CEF-Recall:} Shows how much fraction of OCR tokens in $C_{label}$ is recalled in answer A.
  

The results on our top performing models are shown in \Cref{tab:cef}. We observe that the CEF-Precision score decreases across all models when the summary or the article is given as input compared to the setup where only the OCR text is given. This suggests that the model utilizes less information from the extracted OCR text when it has access to summary or article.  In contrast, CEF-Coverage scores tend to increase when the summary is provided in comparison to without summary as the answer tends to cover more OCR text segments if the summary is present. The CEF-recall scores increase when the summary is provided since many tokens in the summary overlap with the OCR extracted text. 
\Cref{fig:cef_example} illustrates an example of how OCR-extracted texts and the input summary contributed to the answer. }

\subsubsection{Human Evaluation}
\label{human_evalation}
\change{The user interface for the human evaluation annotation task of comparing among generated answers by different models are given in \Cref{human_evaluation_interface}.}

\vspace{5cm}
\subsubsection{Additional Sample Outputs}
\change{Additional sample outputs from the best performing model are shown in \Cref{model_output_samples}.}

\subsubsection{Additional Related Work}
\change{\paragraph{ Query Focused Summarization} In this task, the goal is to generate a summary of the text document(s) based on a given query. For abstractive summarization, a key challenge is that there are only a few small datasets  on specific domains such as news~\cite{dang2005overview} debate~\cite{nema} and meeting~\cite{zhong2021qmsum}.
Given the lack of large training data, some prior methods  \cite{baumel2018query,laskar-etal-2020-wsl,pasunuru2021data} focus on applying transfer learning techniques to first pretrain on a large generic summarization dataset and then finetune on a specific query focused summarization task.
However, these models are not designed to incorporate visual features of charts and their target texts are quite different than ours, as we focus on generating data insights with logical reasoning aspects. }